\documentclass[preprint,10pt,1p,onecolumn]{elsarticle}

\usepackage{fullpage}
\usepackage{times}
\usepackage{hyperref}
\usepackage{amsmath,amssymb,amsfonts}
\usepackage{stfloats}
\usepackage[caption=false,font=footnotesize]{subfig}
\usepackage[hyperref]{xcolor}
\usepackage[algo2e,boxed,ruled,vlined]{algorithm2e} 

\hypersetup{
  colorlinks   = true, 
  urlcolor     = red, 
  linkcolor    = red, 
  citecolor   = red 
}

\usepackage{float}
\usepackage{apalike}
\restylefloat{figure}
\restylefloat{table}

\biboptions{authoryear}

\usepackage{xcolor}

\newcommand{\rev}[1]{#1}
\newcommand{\eaa}[1]{#1}

\newcommand{\eric}[1]{}
\newcommand{\eb}[1]{#1}
\newcommand{\eu}[1]{}

\newcommand{\ecc}[1]{}
\newcommand{\laio}[1]{}

\begin{document}

\begin{frontmatter}

\title{Physics-Informed Neural Nets for Control of Dynamical Systems}


\author[mymainaddress]{Eric Aislan Antonelo }
\author[mymainaddress]{Eduardo Camponogara
}
\ead{eduardo.camponogara@ufsc.br}
\author[mymainaddress,univali]{Laio Oriel Seman} \ead{laio@univali.br}

\author[mymainaddress]{Eduardo Rehbein de Souza
}
\ead{dudurehbein@gmail.com}
\author[mymainaddress]{Jean Panaioti Jordanou
}
\ead{jeanpanaioti@gmail.com}
\author[mymainaddress]{Jomi Fred Hübner
}
\ead{jomi.hubner@ufsc.br}

\cortext[mycorrespondingauthor]{Corresponding author. E-mail: eric.antonelo@ufsc.br}

 
 \address[mymainaddress]{Department of Automation and Systems Engineering, Federal University of Santa Catarina, Florianópolis - Brazil, CEP: 88040-900}
  \address[univali]{Gradute Program in Applied Computer Science, University of Vale do Itajaí, Itajaí - Brazil, CEP: 88302-901}

\begin{abstract}
Physics-informed neural networks (PINNs) impose known physical laws into the learning of deep neural networks, making sure they respect the physics of the process while decreasing the demand of labeled data.  
For systems represented by Ordinary Differential Equations (ODEs),
the conventional PINN has a continuous time input variable and outputs the solution of the corresponding ODE. 
In their original form, PINNs do not allow control inputs, neither can they simulate for \rev{variable} long-range intervals without serious degradation in their predictions.
In this context, this work presents a new framework called \textit{Physics-Informed Neural Nets for Control} (PINC), which proposes a novel PINN-based architecture that is amenable to \emph{control} problems and able to simulate for longer-range time horizons that are not fixed beforehand, 
\eaa{making it a very flexible framework when compared to traditional PINNs. Furthermore, this long-range time simulation of differential equations is faster than numerical methods since it relies only on signal propagation through the network, making it less computationally costly and, thus, a better alternative for simulation of models in Model Predictive Control.
}
We showcase our proposal in the control of two nonlinear dynamic systems: the Van der Pol oscillator and the four-tank system. 
\end{abstract}



\end{frontmatter}


\section{Introduction}


\eaa{
In the era of industry 4.0, 
the simulation and control of complex real-world systems in smart and efficient ways
becomes increasingly important. 
Thus, harnessing deep learning for smart automation and control of real plants is not only desirable but also inevitable. One way to achieve that is by the use of deep neural networks as models in Model Predictive Control (MPC) \citep{Grune:NMPC:2011}.
MPC is a technique that has become standard for multivariate control in industry and academia \citep{Camacho2013}. Since its inception in the 1970s, MPC has been successfully applied in the oil and gas \citep{Jordanou2018}, aerospace \citep{aerospace_mpc} and process industries, as well as in robotics \citep{mpc_robotics}.
The main idea of MPC is to control a system by employing a prediction model: at every iteration of the control loop, an optimization problem is solved using a model of the plant 
in a receding horizon approach.} 

\eaa{
There are two main cases which we consider for which practical application of MPC or even just efficient simulation of a dynamic system are a challenge: (a) sparse or insufficient historical data of the real plant to build a sufficiently accurate machine learning model; (b) when the numerical simulation of a precise model given by Ordinary Differential Equations (ODEs) or Partial Differential Equations (PDEs) is too costly to be considered in a real-time application.
However, a recently introduced approach for training deep neural networks using laws of physics,
namely Physics-Informed Neural Networks (PINN) \citep{Raissi2017,Raissi2019},
is one effective approach that addresses both of the aforementioned challenges.
For the first challenge (a), we assume that a priori knowledge built previously by experts or borrowed from the laws of nature is available.
For (b), instead of relying on numerical solutions of differential equations, PINNs can be used instead easing the computational burden of solving ODEs or PDEs, and consequently extending the application of MPC to more real-time scenarios.
}

A standard PINN has a continuous-time $t$ as input, and the system's state variables as output $\mathbf{y}$.
The main outcome of this approach is that the need for real data collection is reduced to a minimum, since the behavior of the deep network is constrained to follow physical laws given in terms of PDEs or ODEs. These differential equations are included in the learning problem's cost function as nonlinear differential operators on the output $\mathbf{y}$ of the networks, defining a second cost term that regularizes the learning process.
Effectively, this approach also allows solving complex PDEs or ODEs by using deep learning since the network output $\mathbf{y}$ represents the solution of these equations. 
\eaa{
Since \cite{Raissi2017}, many extensions and alternative approaches
have been proposed by \cite{Zhu2019,Sirignano2018,Meng2020,Yang2020,Pang2020,Stinis2020}. 
Applications of PINNs are widespread in many engineering areas, including
\cite{Harp2021} for underground reservoir pressure management, 
\cite{Chen2021} for fatigue data analysis,
and \cite{Kumar2021} for characterizing non-Newtonian fluid flow to achieve optimal use of energy resources.}

However, to the best of our knowledge, there are no PINN architectures 
\eaa{in continuous time}
in the literature that allow optimal control techniques such as Multiple Shooting (MS) \citep{Biegler:SIAM:2010} and  Model-based Predictive Control (MPC) to be readily applied. 
Previous works used neural networks such as Echo State Networks and Long Short-Term Memory (LSTM) networks as models of the plant or process to be controlled \citep{Jordanou2021}. 
These networks are trained exclusively on data collected from the process and are thus not
sample efficient
as PINNs, as the latter can benefit from prior knowledge of the physics laws involved in the processes.
In this sense, the challenge is to make PINNs compliant to control applications so that they can be used as a predictive model of a plant or  process. In their original form, PINNs do not allow control inputs, neither can they simulate for variable long-range intervals without serious degradation in their predictions.

With those limitations in mind, this work presents a new framework called \textit{Physics-Informed Neural Nets for Control} (PINC), which proposes a novel PINN-based architecture
that is amenable to control problems. In particular:
\begin{itemize} 
\item[(i)] 
our PINN-based architecture, called hereafter PINC net, is augmented with extra inputs such as the initial state of the system and control input, in addition to the continuous-time $t$. 
This augmentation is inspired by the multiple shooting and collocation methods \citep{Biegler:SIAM:2010}, which are numerical methods for solving boundary value problems in ODEs, which split the time horizon over which a solution is sought into several shorter intervals (shooting intervals). In our proposal, a single network learns the ODE solution conditioned on the initial state and the given control input 
over the shorter interval.
%
\item[(ii)] 
\rev{this innovation allows enhancing the simulation capabilities of
conventional PINNs, which can not correctly sustain a simulation beyond the time interval that was fixed during network training.
This degradation of the network prediction  is related to the maximum value allowed for $t$, which is fixed at training time.}
On the other hand, our proposed PINC network can run for an indefinite time horizon as long as it is necessary,
without significant deterioration of network prediction. This is achieved by chaining the network prediction in a self-feedback mode, by setting the initial state (input) of the next interval $k$ to the last predicted state (network output) of the previous interval $k-1$.
The work in \cite{Meng2020} also intends to solve this problem, but it requires many individual PINNs to be independently trained 
and, besides, is not ready for control applications.

\item[(iii)] 
the particular structure of PINC makes physics-informed nets \eaa{in continuous time} amenable to MPC applications, which is the first work in the literature to tackle this as far as the authors know.

\item[(iv)] 
\eaa{finally, the real-time requirements for simulation of differential equations, in particular for MPC applications, are better satisfied with PINC than traditional numerical simulation methods, 
since the inference of an already trained PINC network can replace a numerical solution method at each timestep of the prediction horizon in MPC.
}

\end{itemize}

In the following, some related works are presented in Section~\ref{sec:relatedworks}. PINNs, MPC and PINC are introduced in Section \ref{sec:methods}, whereas the experiments on identifying and controlling two known dynamical systems in the literature are shown in Section~\ref{sec:experiments}. Section~\ref{sec:conclusion} concludes this work.

\section{Related Works}
\label{sec:relatedworks}

\subsection{Neural networks and MPC}
\eaa{
Neural networks have long been used as models in MPC tasks or as controllers themselves. 
Previous works have trained neural networks to imitate MPC strategies using the usual mean squared error cost functions \citep{Ortega1996,Cavagnari1999}.
In \cite{Aakesson2006}, the control law is represented by a neural network approximator, trained offline to minimize a control-related cost function directly, without the need to calculate a model predictive controller during training. 
}

\eaa{
In the vein of Recurrent Neural Networks (RNN), works such as \cite{Jordanou2021} and \cite{echo_pred} utilize Echo State Networks as dynamical models for the MPC.
\cite{Jordanou2021} uses a Trajectory linearization approach \citep{Maciej2014}, by derivating the input-output sensitivities along the nonlinear free response over the prediction horizon to calculate a forced response \citep{Camacho2013}.
In \cite{echo_pred}, the whole ESN is approximated into a state space system for computation of the control action.
\cite{Terzi2020} rely on the same reduction approach, however employing LSTMs instead of ESN.
}

\eaa{
Another example is the classical Approximate Predictive Control \citep{APC}, which employs a feedforward neural network that implements dynamics through the application of delayed outputs as inputs (an external dynamics model \citep{nl_sys_ident}), obtains an ARX 
(Auto Regressive with eXogenous input) model from the networks through derivation, and performs GPC (Generalized Predictive Control) calculations per time step \citep{Camacho2013}.
\cite{Hertneck2018} consider a neural network as the approximation to a MPC, in the same vein as \cite{Aakesson2006}.
}

\subsection{Long-range simulation with PINNs}
\eaa{
In \cite{Meng2020}, \textit{parareal} PINNs are proposed for long time integration of time-dependent PDEs. 
They decompose a long-time problem in several short-time independent problems supervised by a fast coarse-grained solver, which provide approximate predictions of the solution at discrete times.
Several smaller, fine PINNs are trained in parallel with the help of the supervision given by the fast solver. Each PINN solves the problem for a particular time interval independently.
Notice that their approach does not include the possibility of control inputs and, thus, can not be readily used for control applications.
On the other hand, while the initial goal of our proposal is to extend PINNs for control, we also benefit from being able to simulate a PINN for ODEs for a long time interval as well (we expect that our architecture can be extended for PDEs too). 
}

\subsection{PINNs for control}
\eaa{Since the first appearance of this work \citep{Antonelo2021}, some architectures based on PINNs for control have been introduced.
In \cite{Zhai2021}, PINNs are supposedly used to control chaos in van der Pol oscillating circuits. However, the circuits employed in their paper have no control input, neither their method outputs a control signal to be applied. Their PINN architecture still has only time $t$ as input, and has an unusual loss function which includes the loss for the data points and also for the reference to be followed. Surprisingly, we have not found any physics law that was included in the network training, making their network just an ordinary one.
Besides, no control input can be derived from it to actually control a plant or dynamical system.
}

\eaa{
In \cite{Liu2021}, a model-based Reinforcement Learning (RL) algorithm for the first time employs physical laws to learn the state transition dynamics of an agent's environment. The model corresponds to an encoder-decoder recurrent network architecture that learns the state transition function by minimizing the violation of conservation laws. The real samples (state-action data pairs and corresponding rewards) from the environment are used to train the agent and the transition model simultaneously.
In turn, the latter is used to generate samples into an 
alternative replay buffer that ultimately improves sample efficiency in the RL update and reduces real-world interaction. 
As the transition function is part of a Markov Decision Process (MDP) formulation, it represents a discrete evolution of the environment dynamics. For this reason, the physical loss function is built on the laws of the system in their discretized form, instead of the continuous form as proposed in our work. Notice that while they require training a recurrent network, our work is based on feedforward networks as time is explicitly given as input here. 
}

\eaa{
In \cite{Gokhale2022}, PINNs are employed to learn a control-oriented thermal model of a building. 
As in \cite{Liu2021}, they assume that the model is a discrete transition function in a MDP that predicts the next state, given the current state and action. In that way, control actions could be input to the model. Their physical loss also has to be discretized, differently from our work.
Although their proposal is control-oriented, they do not show actual control experiments with the trained PINNs, as we do in Section~\ref{sec:experiments}.
}

\section{Methods}
\label{sec:methods}

\subsection{Physics-informed Neural Networks (PINNs)} \label{sec:PINN}

\cite{Raissi2017,Raissi2019} introduced physics-informed neural networks, training deep neural networks in a supervised way to respect any physical law described by partial differential equations (PDEs). 
The PINN approach allows one to find data-driven solutions of PDEs or ODEs automatically.
In this paper, nonlinear ODEs are considered in the following general form:
\begin{equation}
   \partial_t \mathbf{y} + \mathcal{N}[\mathbf{y}]=0,  \quad t \in [0,T] \label{eq:general_ode}
\end{equation}
where $\mathcal{N}[\cdot]$ is a nonlinear differential operator and $\mathbf{y}$ represents the state of the dynamic system (the latent ODE solution).

We define $\mathcal{F}(\mathbf{y})$ to be equivalent to the left-hand side of Equation (\ref{eq:general_ode}):
\begin{equation}
\mathcal{F}(\mathbf{y})  := \partial_t \mathbf{y} + \mathcal{N}[\mathbf{y}] 
\label{eq:ode}
\end{equation}

Here, $\mathbf{y}$ also represents the output of a multilayer neural network (hence the notation $\mathbf{y}$ instead of $\mathbf{x}$) which has the continuous time $t$ as input:
$\mathbf{y} = f_\mathbf{w}(t)$, where $f_\mathbf{w}$ represents the mapping function obtained by a deep  network parameterized by adaptive weights $\mathbf{w}$.
This formulation implies that a neural network must learn to compute the solution of a given ODE.

Assuming an autonomous system for this formulation, a given neural network $\mathbf{y}(t)$ is trained using optimizers such as ADAM \citep{Kingma2014} or L-BFGS \citep{Andrew2007} to minimize
a mean squared error (MSE) cost function:
\begin{equation}
\textrm{MSE} = \textrm{MSE}_y + \textrm{MSE}_{\mathcal{F}},
\label{eq:erro}
\end{equation}
where
\begin{subequations}
\begin{align}
\textrm{MSE}_y = \frac{1}{N_y} \sum_{i=1}^{N_y}  \frac{1}{N_t} \sum_{j=1}^{N_t} | y_i(t^j) - \widehat{y}_i^j |^2,
\label{eq:erro1}
\end{align}
\begin{align}
\textrm{MSE}_{\mathcal{F}} =  \frac{1}{N_y} \sum_{i=1}^{N_y} 
   \frac{1}{N_{\mathcal{F}}} \sum_{k=1}^{N_{\mathcal{F}}}  | \mathcal{F}( y_i(t^k) ) |^2,
\label{eq:erro2}
\end{align}
\end{subequations}
where: $N_t$, $N_{\mathcal{F}}$, and $N_y$ correspond to the number of training data samples, the number of collocation points, and the number of outputs of the neural network, respectively; $y_i(\cdot)$ is the $i$-th output of the network; ${\widehat{y}_i^j}$ represents the desired $i$-th output for $y_i(\cdot)$, considering the $j$-th data pair $(t^j,\widehat{y}_i^j)$.
The first loss term $\text{MSE}_y$ corresponds to the usual cost function for regression \citep{Bishop2006}
based on collected training data 
$\{(t^j,\widehat{y}_i^j)\}_{j=1}^{N_t}$, which usually provides the boundary (initial or terminal) conditions of ODEs when solving these equations.

The second loss term $\textrm{MSE}_{\mathcal{F}}$ penalizes the misadjusted behavior of $\mathbf{y}(t)$,
measured by $\mathcal{F}(\mathbf{y})$ in Equation (\ref{eq:ode}), whereby the physical structure of the solution is imposed by $\mathcal{F}(\mathbf{y})$ at a finite set of randomly sampled collocation points
$\{t^k\}_{k=1}^{N_{\mathcal{F}}}$.
Experiments show that the training data size $N_t$ required for learning a certain dynamical behavior
is drastically reduced due to the a priori information assimilated from $\textrm{MSE}_{\mathcal{F}}$.
As the differential equation of the physical system is assumed to be represented by $\mathcal{F}(\mathbf{y}) = 0$, the term $\textrm{MSE}_{\mathcal{F}}$ is a measure of how well the PINN adheres to the solution of the physical model.
This physics-informed approach provides a framework that unifies a previously available theoretical, possibly approximate model and measured data from processes, which is capable of correcting imprecisions in the theoretical model or providing sample efficiency in process modeling.

\subsection{Nonlinear Model Predictive Control} \label{sec:MPC}

Model Predictive Control (MPC) has evolved considerably over the last two decades, significantly impacting industrial process control. This impact can be attributed to its generality in posing the process control problem in the time domain, being suitable for SISO (Single-Input Single-Output), and MIMO (Multiple-Input Multiple-Output) systems. Soft and hard constraints can be imposed on the formulation of the control law through optimization problems, while minimizing an objective function over a prediction horizon \citep{Normey-Rico2007}.

\begin{figure}
	\centering
	\includegraphics[width=0.7\columnwidth]{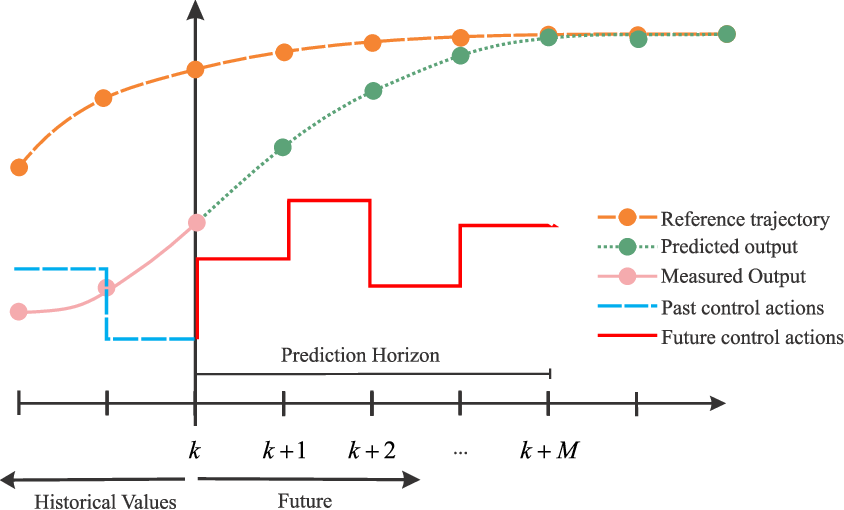}
	\caption{Representation of the output prediction at a time instant $t_k$, where the proposed actions generate a predicted behavior that reduces the distance between the value predicted by the model and a reference trajectory.
	}
	\label{fig:mpc_pred}
\end{figure}

MPC is not a specific control strategy, but rather a denomination of a vast set of control methods developed considering some standard ideas and predictions \citep{Normey-Rico2007}. Figure \ref{fig:mpc_pred} shows a representation of the output prediction at a time instant, where the proposed actions generate a predicted behavior that reduces the distance between the value predicted by the model and a reference trajectory.

The MPC strategy uses a discrete mathematical model based on the real process of interest. A predicted output is calculated in a prediction horizon by comparing the mathematical model to the real process's output. 
 To propose control actions, the MPC strategy uses an iterative optimization process, taking into account the mathematical model of interest and the constraints that it is subjected. Based on objectives and constraints, the optimization problem is composed of mathematical expressions established in the controller's design phase, taking many forms. Usually, quadratic functions are used to penalize the error in the reference tracking.

According to \cite{Camacho2013},
there are several ways to classify these controllers taking into account characteristics such as model linearity, treatment of uncertainties, and how the optimization problem is solved. In this work, we focus on the lack of model linearity, more specifically in the Nonlinear Model Predictive Control (NMPC) \citep{Grune:NMPC:2011}. The discrete NMPC formulation is given by:
\begin{small}
\begin{subequations}
\begin{equation}
    J=\sum_{j=N_1}^{N_2}\left\lVert \mathbf{x}[k+j]-\mathbf{x}^{\rm ref}[k+j] \right\rVert_{\mathbf{Q}}^2  +\sum_{i=0}^{N_u-1}\left\lVert \Delta \mathbf{u}[k+i]\right\rVert_{\mathbf{R}}^2 \label{eq:mpc_cost}
\end{equation}
while being subject to:
\begin{align}
    & \mathbf{x}[k+j+1] = \mathbf{f}(\mathbf{x}[k+j],\mathbf{u}[k+j]), \quad  \forall j = 0,\ldots,N_2-1 \label{eq:state_const} \\
    & \mathbf{u}[k+j] = \mathbf{u}[k-1] + \sum_{i=0}^{j}\mathbf{\Delta u}[k+i], \quad   \forall j=0,\ldots,(N_u - 1) \label{eq:ctrl_action} \\
    & \mathbf{u}[k+j] = \mathbf{u}[k+N_u - 1], \quad \forall j=N_u,\ldots,N_2-1 \label{eq:ctrl_action_B}\\
    & \mathbf{h}(\mathbf{x}[k+j],\mathbf{u}[k+j]) \leq 0, \quad \forall j = N_1,\ldots,N_2  \label{eq:ineq_const}\\
    & \mathbf{g}(\mathbf{x}[k+j],\mathbf{u}[k+j]) = 0, \quad \forall j = N_1,\ldots,N_2  \label{eq:eq_const}
\end{align}
\end{subequations}
\end{small}
where $k$ represents the time step at which the MPC problem is being computed, $\mathbf{x}[k]$ is the recurrent state of the dynamic system which, for simplification purposes, is also the output (i.e., $\mathbf{x}=\mathbf{y}$), $\mathbf{x}^{\rm ref}$ is the set-point signal over the prediction horizon (i.e., reference), being defined by the first penalized instant $k+N_1$ and the last instant $k+N_2$. The cost function $J$ is the penalization of the quadratic error between the model output $\mathbf{x}$ and the reference $\mathbf{x}^{\rm ref}$ along the horizon, and the penalization of the control increment $\mathbf{\Delta u}$.
Each penalization is weighted by the diagonal matrices $\mathbf{Q}$ and $\mathbf{R}$, respectively.
Eqn. \eqref{eq:state_const} is the constraint imposed by the considered state-equation model with $\mathbf{x}$ as the state, and equations \eqref{eq:ineq_const} and \eqref{eq:eq_const} refer to inequality and equality constraints imposed by functions $\mathbf{h}$ and $\mathbf{g}$, respectively.
Eqns. \eqref{eq:ctrl_action} and \eqref{eq:ctrl_action_B} define the relation between the control action $\mathbf{u}$ and the control increments, which are aggregated into the control action from time $k$ up until either the control action time $k+j$ or $k+ N_u - 1$.

The optimization problem is defined by equations from \eqref{eq:mpc_cost} to \eqref{eq:eq_const} and results in a Non-Linear Programming (NLP) Problem, which can be solved using well-established methods like Sequential Quadratic Programming (SQP) \citep{Nocedal2006} and the Interior-Point (IP) method, available in commercial \citep{Gill:SNOPT:2005} and non-commercial solvers \citep{Wachter2006}. The NLP is solved at each time step $k$, and typical approaches only apply the first control increment into the system  \citep{Camacho2013}.
%

\subsection{Physics-Informed Neural nets-based Control (PINC)} \label{sec:PINC}

\begin{figure}[tb]
	\centering
	\includegraphics[width=0.5\linewidth]{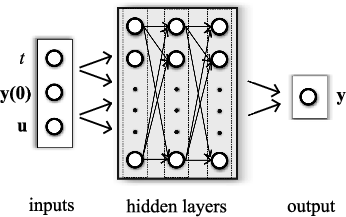}
	\caption{The PINC network has initial state $\mathbf{y}(0)$ of the dynamic system and control input $\mathbf{u}$ as inputs, in addition to continuous time scalar $t$. Both  $\mathbf{y}(0)$ and $\mathbf{u}$ can be multidimensional. The output $\mathbf{y}(t)$ corresponds to the state of the dynamic system as a function of $t \in [0,T]$, and initial conditions given by $\mathbf{y}(0)$ and $\mathbf{u}$. The deep network is fully connected even though not all connections are shown.
	}
	\label{fig:pincnet}
\end{figure}

Unlike PINNs that assume fixed inputs and conditions, the proposed PINC framework operates with variable initial conditions as well as control inputs that can change over the complete simulation, making it suitable for model predictive control tasks. The network is augmented with two, possibly multidimensional inputs: control action $\mathbf{u}$ and initial state $\mathbf{y}(0)$, as illustrated in Figure~\ref{fig:pincnet}.
The output of the network is given by:
\begin{equation}
\mathbf{y}(t)=
f_\mathbf{w}(t,\mathbf{y}(0),\mathbf{u}), 
\quad t \in [0,T]
\label{eq:net_pred}
\end{equation}
where $f_\mathbf{w}$ represents the mapping given by a deep network parameterized by weights $\mathbf{w}$.
In this work, we assume the control input to be a constant value for the time interval $t \in [0,T]$. 
Thus, the new formulation provides a conditioned response $\mathbf{y}(t)$ on $\mathbf{u}$ and $\mathbf{y}(0)$
during this interval of $T$ seconds.

Traditional PINNs tend to degrade rapidly for long time intervals and can only accept input $t$ in the range the network was trained.
The PINC framework significantly alleviates this degrading issue as well as enables control applications by
dividing the problem in $M$ equidistant  control time intervals, each of $T$ seconds
(see Fig.~\ref{fig:pincevolution}).
We call this shorter period of $T$ seconds as the \textit{inner continuous time interval} of the problem, in which a solution of an ODE is obtained given some initial condition 
$\mathbf{y}(0)$  (which models the current system state) and  control input $\mathbf{u}$ for $t \in [0,T]$.
This ODE solution $\mathbf{y}(t)$, which is the output of the network, is found by a single PINC network, that is, the same network solves all $M$ intermediate problems, which results from learning the ODE solution for a particular range of initial conditions and control inputs that vary over the complete time horizon, but which stay constant for $t \in [0,T]$.

\subsubsection{Combining the intermediate solutions}

Each of the $M$ intermediate solutions of $T$ seconds can be viewed in Figure~\ref{fig:pincevolution}. The states $\mathbf{y}[k]$ inferred by the network can be seen at the top of the figure as a dashed trajectory, while its corresponding inputs are located in the lower part. Here, the notation changed to represent the output in discrete time $k$. Between steps $k$ and $k+1$, one intermediate solution is given by Equation (\ref{eq:net_pred}), fixing the control input to some constant and the initial state to the last state of step $k-1$.

Since $t$ is an input to the network, the state at $t=T$ can be directly inferred by a single forward network propagation:
\begin{equation}
   \mathbf{y}[k]=f_\mathbf{w}(T,\mathbf{y}[k-1], \mathbf{u}[k]) \label{eq:recurrent_pred}
\end{equation}
where the initial state is set to the last state of the previous step, i.e.,  $\mathbf{y}[k-1]$; and the control input $\mathbf{u}[k]$ has an index $k$ indicating which fixed value is applied in the inner continuous time interval between steps $k-1$ and $k$.

\subsubsection{Free-run simulation in the prediction horizon}

The initial state of the dynamical system in step $k$ can be either the true state
$\mathbf{\widehat{y}}[k-1]$ coming from the process
or the previous network prediction $\mathbf{y}[k-1]$ at timestep $k-1$.

\begin{figure}[tb]
	\centering
	\subfloat[PINC in self-loop mode]{
	\includegraphics[width=0.37\linewidth]{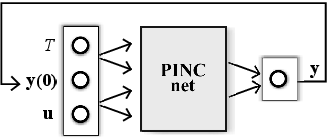}	
	\label{fig:pinc_feedback}
	}
	\qquad
	\subfloat[PINC connected to the plant]{
	\includegraphics[width=0.37\linewidth]{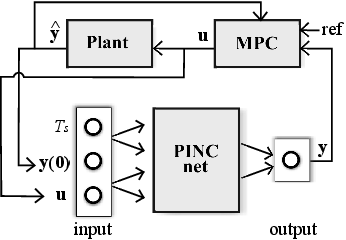}
	\label{fig:pinc_plant}
	}
	\caption{
	Modes of operation of the PINC network.
	(a) PINC net operates in self-loop mode, using its own output prediction as next initial state, after $T$ seconds. This operation mode is used within one iteration of MPC, for trajectory generation until the prediction horizon of MPC completes 
	(predicted output from Fig.~\ref{fig:mpc_pred}).
	(b) Block diagram for PINC connected to the plant. One pass through the diagram arrows corresponds to one MPC iteration applying a control input $\mathbf{u}$ for $T_s$ timesteps for both plant and PINC network. Note that the initial state of the PINC net is set to the real output of the plant. In practice, in MPC, these two operation modes are executed in an alternated way (optimization in the prediction horizon, and application of control action).
	}
	\label{fig:pinc_modes}
\end{figure}

Within one iteration of MPC, the PINC net is used for a certain prediction horizon without feedback from the process.
This means that the network prediction $\mathbf{y}[k-1]$ and not the true state $\mathbf{\widehat{y}}[k-1]$ 
is fed back as input to the same network in the next timestep $k$ of the prediction horizon (Fig.~\ref{fig:pinc_feedback}).

In discrete time control applications, a sampling period $T_s$ must be chosen.
The setting of $T_s$ usually depends on the particular dynamics of the process being modeled.
Here, $T$ is equal to the sampling period $T_s$. Thus, using Equation \eqref{eq:recurrent_pred}, we can encapsulate the PINC prediction function so that it is only a function of the control action $\mathbf{u}[k]$ and previous prediction $\mathbf{y}[k-1]$, leaving $T$ implicit:
\begin{align}
  \mathbf{y}[k] & = \widehat{f}_\mathbf{w}(
  \mathbf{y}[k-1], \mathbf{u}[k] ) \nonumber   \\
   &  = f_\mathbf{w}(T,\mathbf{y}[k-1], \mathbf{u}[k])  \label{eq:interface}
\end{align}

\begin{figure}
	\centering
	\includegraphics[width=0.75\linewidth]{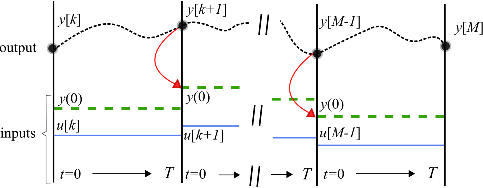}
	\caption{Representation of a trained PINC network evolving through time in self-loop mode (Fig.~\ref{fig:pinc_feedback}) for trajectory generation in prediction horizon. 
	The top dashed black curve corresponds to a predicted trajectory $\mathbf{y}$ of a hypothetical dynamic system in continuous time. 
	The states $\mathbf{y}[k]$ are snapshots of the system in discrete time $k$ positioned at the equidistant vertical lines. 
Between two vertical lines (during the \textit{inner continuous interval} between steps $k$ and $k+1$), the PINC net learns the solution of an ODE with $t \in [0,T]$, conditioned on a fixed control input $\mathbf{u}[k]$ (blue solid line) and initial state $\mathbf{y}(0)$ (green thick dashed line).
Control action $\mathbf{u}[k]$ is changed at the vertical lines and kept fixed for $T$ seconds, and the initial state $\mathbf{y}(0)$ in the interval between steps $k$ and $k+1$ is updated to the last state of the previous interval $k-1$ (indicated by the red curved arrow).
The PINC net can directly predict the states at the vertical lines without the need to infer intermediate states $t < T$ as numerical simulation does.
	 Here, we assume that $T=T_s$ and, thus, the number of discrete timesteps $M$ is equal to the length of the prediction horizon in MPC.
} \label{fig:pincevolution}
\end{figure}

We call $\widehat{f}_\mathbf{w}$ the control interface for the PINC framework. Thus, $ \frac{\partial \widehat{f}_\mathbf{w}}{\partial\mathbf{u}}$ can be computed for providing the Jacobian matrix to solvers used in MPC, possibly by means of automatic differentiation.
  This control interface provides the prediction of the states of the dynamic system at the vertical lines in Fig.~\ref{fig:pincevolution}, that is, at every $T_s$ seconds, the state $\mathbf{y}[k]$ is predicted in a single forward net propagation operation, for $k=1,...,M$. This differs from numerical integration methods that need to integrate over the continuous inner interval \citep{Iserles:Book:1996}.


Since the prediction is fed back as an input at every discrete timestep, it is expected that errors accumulate in the long free run. This is not exclusive of this approach, and is common to recurrent neural networks.
However, because MPC works in a receding horizon control approach, at every timestep $k$ of the control loop, the input $\mathbf{y}[k-1]$ representing the initial state  is set to the real system's state $\mathbf{\widehat{y}}[k-1]$ (Fig. \ref{fig:pinc_plant}). 
Thus, the prediction horizon in MPC always starts from the true initial state $\mathbf{\widehat{y}}[k-1]$, that is, Equation (\ref{eq:interface}) becomes 
\begin{equation}  \mathbf{y}[k]  = \widehat{f}_\mathbf{w}(
  \mathbf{\widehat{y}}[k-1], \mathbf{u}[k]
  )
\end{equation}
which counters error accumulation between consecutive control iterations.

The error might accumulate when the MPC model is used in a future finite prediction horizon
to solve a constrained optimization problem. In this case, the prediction $\mathbf{y}[k-1]$ is fed back as no readings from the real process at a future time are possible (Fig.~\ref{fig:pinc_feedback}).

\subsubsection{Training}
\label{sec:pinc_training}

The first loss term in Equation (\ref{eq:erro}) can be generalized to the PINC network as:
\begin{equation}
   \textrm{MSE}_y = \frac{1}{N_y} \sum_{i=1}^{N_y}  \frac{1}{N_t} \sum_{j=1}^{N_t} | y_i(\mathbf{v}^j) - \widehat{y}_i^j |^2,
\label{eq:mse_y_pinc}
\end{equation}
where the pair $(\mathbf{v}^j, \widehat{\mathbf{y}}^j)$ corresponds to the $j$-th training example and $\mathbf{v}^j=(t,\mathbf{y}(0),\mathbf{u})^j$ is the whole input to the network (i.e., time, initial state, and control input).
Usually, this dataset comes from measured data, but in this work we will show that, if we assume that the given ODE is an exact representation of the process, it is enough for this dataset to contain only the initial conditions of the modelled ODE.
   For instance, one such training data pair is $((0,0.4,0.6), 0.4)$, which means: at $t=0$ the initial state is 0.4, the control input is 0.6 and the desired output is equal to the initial state (0.4).
Thus, for all data points, $t=0$, while $\mathbf{y}(0)$ and $\mathbf{u}$ are randomly sampled from intervals defined according to the modelled dynamic system. %
   This means that $\text{MSE}_y$ represents the mean squared error for all randomly sampled initial conditions of the considered ODE and control inputs.
Note also that  the input $\mathbf{y}^j(0)$ is equal to the desired output $\widehat{\mathbf{y}}^j$ in the training set, such that the network must learn to \textit{reproduce} the initial state $\mathbf{y}^j(0)$ into the network output $\mathbf{y}(\mathbf{v}^j) $ at $t=0$.
In practice, the aforementioned assumption allows training using  randomly sampled data for solving the ODE, without ever requiring measured process's data. 

The second loss term in Equation (\ref{eq:erro}) is rewritten as:
\begin{equation}
  \textrm{MSE}_{\mathcal{F}} =  \frac{1}{N_y} \sum_{i=1}^{N_y} 
    \frac{1}{N_{\mathcal{F}}} \sum_{k=1}^{N_{\mathcal{F}}}  | \mathcal{F}( y_i(\mathbf{v}^k) ) |^2,
\label{eq:mse_F_pinc}
\end{equation}
where $\mathbf{v}^k$ corresponds to the $k$-th collocation point $(t,\mathbf{y}(0),\mathbf{u})^k$, where now all three types of inputs (and not only the last two), i.e., time, initial condition, and control input, are randomly sampled from their respective particular intervals. 
   Specifically, the interval for $t$ is $[0,T]$, where $T$ is the \textit{inner continuous interval} of the PINC framework. 

Basically, this formulation means that the PINC net is trained with \textit{data} points that lie on the boundary of simulations, i.e., only initial states of ODEs are presented for the loss function in Eq. \eqref{eq:mse_y_pinc}. Practically, this does not require collecting data from ODE simulators.
On the other hand, the collocation points in $\textrm{MSE}_{\mathcal{F}}$ serve to regularize the PINC net to satisfy the behavior defined by $\mathcal{F}$. 
Thus, in the training process, the PINC net is only directly informed with a initial state in Eq. \eqref{eq:mse_y_pinc}, and its physics-informed cost loss in Eq. \eqref{eq:mse_F_pinc} must enforce the structure of the differential equation into its output $\mathbf{y}(\cdot)$ for the remaining \textit{inner continuous interval} of $T$ seconds (e.g., $t \in (0,T]$).

The total loss can be generalized to
$\textrm{MSE} = \textrm{MSE}_y + \lambda \cdot \textrm{MSE}_{\mathcal{F}}$, where $\lambda$ represents a rescaling factor so that both terms are approximately in the same scale.
Once the PINC net structure, datasets and the losses are defined, the training process starts with the ADAM optimizer \citep{Kingma2014} for $K_{1}$ epochs, and subsequently continues with the L-BFGS optimizer \citep{Andrew2007} for $K_{2}$ iterations in order to adapt the net weights $\mathbf{w}$ towards the minimization of \textrm{MSE}. 
Note that automatic differentiation is employed for the physics-informed term $\textrm{MSE}_{\mathcal{F}}$ in Eq. \eqref{eq:mse_F_pinc}, using deep learning frameworks such as \textit{Tensorflow}.

\subsubsection{NMPC}
\label{sec:pinc_nmpc}

After training, the PINC net is used as a model in nonlinear MPC, whose algorithm is described in Section \ref{sec:MPC}.
Thus, the control interface function $\widehat{f}_{\mathbf{w}}$ in Equation (\ref{eq:interface}) replaces Equation (\ref{eq:state_const}) in the MPC formulation, redefining the notation of a dynamic system's state by the prediction given by the PINC network, i.e., $\mathbf{x}[k] = \mathbf{y}[k]$.
After these substitutions, we arrive at a 
Multiple Shooting (MS)-inspired formulation 
for the NMPC problem under the PINC framework:
\begin{subequations}
\begin{equation}
    J=\sum_{j=N_1}^{N_2}\left\lVert \mathbf{y}[k+j]-\mathbf{y}^{\rm ref}[k+j] \right\rVert_{\mathbf{Q}}^2  +\sum_{i=0}^{N_u-1}\left\lVert \Delta \mathbf{u}[k+i]\right\rVert_{\mathbf{R}}^2 \label{eq:MS:cost}
\end{equation}
while being subject to:
\begin{align}
    & \mathbf{y}[k+j+1] = \widehat{\mathbf{f}}_{\mathbf{w}}(\mathbf{y}[k+j],\mathbf{u}[k+j]), \quad \forall j = 0,\ldots,N_2-1 \label{eq:MS:state_const} \\
  &  \mathbf{u}[k+j] = \mathbf{u}[k-1] + \sum_{i=0}^{j}\mathbf{\Delta u}[k+i], \quad \forall j=0,\ldots,(N_u - 1) \label{eq:MS:ctrl_action} \\
    & \mathbf{u}[k+j] = \mathbf{u}[k+N_u - 1], \quad \forall j=N_u,\ldots,N_2-1 \label{eq:MS:ctrl_action_B}\\
    & \mathbf{h}(\mathbf{y}[k+j],\mathbf{u}[k+j]) \leq 0, \quad \forall j = N_1,\ldots,N_2  \label{eq:MS:ineq_const}\\
    & \mathbf{g}(\mathbf{y}[k+j],\mathbf{u}[k+j]) = 0, \quad \forall j = N_1,\ldots,N_2  \label{eq:MS:eq_const}
\end{align}
\end{subequations}

\subsection{Metrics}
%
The evaluation of the PINC net prediction performance is done on a validation set in self-loop mode (Figures~\ref{fig:pinc_feedback} and \ref{fig:pincevolution}). 
In particular, the generalization MSE is computed only at the discrete time steps (vertical lines in Fig.~\ref{fig:pincevolution}):
\begin{equation}
   \textrm{MSE}_{gen} = \frac{1}{N_y} \sum_{i=1}^{N_y} \frac{1}{N} \sum_{k=1}^{N} \bigl ( 
y_i[k] - \widehat{y_i}[k] \bigr )^2,
\label{eq:mse_gen}
\end{equation}
where: $\mathbf{y}[k]$ is the prediction of the PINC net given by Equation \eqref{eq:interface} and $\mathbf{\widehat{y}}[k]$ is obtained with Runge-Kutta (RK) simulation of the true model of the plant; 
\eaa{$N$ is the length of vector 
$\mathbf{y}$; }
and the same control input signal $\mathbf{u}[k]$ is given to both the PINC net and the RK model.

The control performance is measured by employing the \textbf{Integral of Absolute Error (IAE)} on a simulation of $C$ iterations: 
\begin{equation}
    \textrm{IAE} = 
    \frac{1}{N_y} \sum_{i=1}^{N_y} \sum_{k=1}^{C} \bigl 
    |{y^{\rm ref}_i}[k] - y_i[k] \bigr |
    \label{eq:iae_prediction}
\end{equation}
and the \textbf{Root Mean Squared Error (RMSE)}:
\begin{equation}
    \textrm{RMSE} =
    \frac{1}{N_y} \sum_{i=1}^{N_y}
    \sqrt{\frac{1}{C}\sum_{k=1}^C\bigl ( y^{\rm ref}_i[k] - y_i[k]\bigl )^2}
\end{equation}
where $y^{\rm ref}_i[k]$ is the reference value of $y_i[k]$ at timestep $k$.

The IAE is ideal for comparing simulation runs with the exact same reference signal, as the sum of absolute errors is very sensitive to changes in control performance \citep{iae}.
Meanwhile, the RMSE can capture the average error behavior of the controller.


\subsection{PINC Algorithms}


\eaa{In this section, an overview of the proposal is presented with the help of high-level algorithms.
Algorithm \ref{algor:pinc}'s objective is training the PINC network. It uses datapoints and collocations points (generated as described in Section~\ref{sec:pinc_training}) to minimize
Eq. \eqref{eq:mse_y_pinc} + Eq. \eqref{eq:mse_F_pinc}, first with ADAM optimizer and then with L-BFGS optimizer.
}

\begin{algorithm2e}
\SetKwInput{KwIn}{input}
\SetKwInput{KwOut}{output}
\KwIn{
$K_1$,
$K_2$,
$\mathcal{F(\cdot)}$,
$\{(\mathbf{v}^j, \widehat{\mathbf{y}}^j):j=1,\dots,N_t\}$,
$\{\mathbf{v}^k:k=1,\dots,N_{\mathcal{F}}\}$;
}

{\bf initialize} PINC weights $\mathbf{w}$ with Xavier normal distribution;\\

\tcp{Train with ADAM}
\For{$K_1$ \emph{epochs}}{

  Compute the gradients of Eq. \eqref{eq:mse_y_pinc} + Eq. \eqref{eq:mse_F_pinc} (with $\mathcal{F(\cdot)}$) with respect to 
  $\mathbf{w}$ using the data points
  $\{(\mathbf{v}^j, \widehat{\mathbf{y}}^j):j=1,\dots,N_t\}$ 
  and collocation points
$\{\mathbf{v}^k:k=1,\dots,N_{\mathcal{F}}\}$;
  \\
  Update $\mathbf{w}$ with ADAM optimizer and the obtained gradients;

}

\tcp{Train with L-BFGS}
\For{$K_2$ \emph{iterations}}{

    Compute the gradients of Eq. \eqref{eq:mse_y_pinc} + Eq. \eqref{eq:mse_F_pinc} (with $\mathcal{F(\cdot)}$)  with respect to 
  $\mathbf{w}$ using the data points
  $\{(\mathbf{v}^j, \widehat{\mathbf{y}}^j):j=1,\dots,N_t\}$ 
  and collocation points
$\{\mathbf{v}^k:k=1,\dots,N_{\mathcal{F}}\}$;
  \\
  Update $\mathbf{w}$ with L-BFGS optimizer and the obtained gradients;
    \\
  Save network $\mathbf{w}$ with best performance seen so far on a validation set using Eq. \eqref{eq:mse_gen};

}
\KwOut{network $\mathbf{w}$ with lowest validation error;}
\caption{PINC Training Algorithm}
\label{algor:pinc}
\end{algorithm2e}


\eaa{Algorithm \ref{algor:mpc} employs MPC with PINC using the minimization process (NMPC) described in the Section \ref{sec:pinc_nmpc} for each timestep $k$ out of $C$ iterations (i.e., total length of the reference signal) to yield a control action $\mathbf{u}[k]$ to be applied to the plant.
}

\begin{algorithm2e}
\SetKwInput{KwIn}{input}
\SetKwInput{KwOut}{output}
\KwIn{$\mathbf{Q}$, $\mathbf{R}$, $N_u$, $N_1$, $N_2$, $\mathbf{y}^{\rm ref}$,
$\widehat{\mathbf{f}}_{\mathbf{w}}$, $\mathbf{x}[0]$}

\tcp{Use the trained PINC to perform the control procedure}

\For{$k:=0,1,2,\ldots,C$}{
  Set initial state $\mathbf{y}[k]$ to the plant's current state $\mathbf{x}[k]$;
  \\
  Minimize \eqref{eq:MS:cost} s.t. \eqref{eq:MS:state_const}---\eqref{eq:MS:eq_const}, with respect to control $\mathbf{u}$  for reference $\mathbf{y}^{\tt ref}$ at timestep $k$,
  using the trained network 
  $\widehat{\mathbf{f}}_{\mathbf{w}}$ as predictive model, control horizon $N_u$, prediction horizon $M=N_2-N_1+1$, 
  and weight matrices $\mathbf{Q}$ and $\mathbf{R}$;
  \\
  Apply $\mathbf{u}[k]$ to the plant, obtaining the next states $\mathbf{x}[k+1]$;
}

\KwOut{control action $\mathbf{u}$}
\caption{MPC with PINC Algorithm}
\label{algor:mpc}
\end{algorithm2e}

\section{Experiments} \label{sec:experiments}

This section presents experiments regarding the application of PINC to the modeling and control of the Van der Pol Oscillator and the four-tank system, which are two dynamical systems often considered for nonlinear analysis in the literature.

\subsection{Van der Pol Oscillator}

\subsubsection{Model}
The Van der Pol oscillator \citep{hafeez2015} is an ODE initially discovered by Balthazar Van der Pol that had the original purpose of modeling triode oscillations in electric circuits.
Since then, the ODE has been used for other purposes, such as seismology and biological neuron modeling \citep{hafeez2015}, and as a standard proof-of-concept dynamical system for optimal control applications \citep{andersson2012}.
The equations that govern the Van der Pol Oscillator are as follows:
\begin{subequations}
\begin{align}
    \dot{x}_1 &= x_2\\
    \dot{x}_2 &= \mu(1 - x_1^2)x_2 - x_1 + u
\end{align}
\end{subequations}
where $\mu = 1$ is referred to as the damping parameter, which affects how much the system will oscillate, $\mathbf{x} = (x_1,x_2)$ is the system state, and $u$ is an exogenous control action.

By inspection, the Van der Pol oscillator has an equilibrium at $\bar{\mathbf{x}} = (u,0)$, which is stable for  a constant input  $u\in (-\sqrt{3},-1)$ or $u\in (1,\sqrt{3})$. 
The oscillator also has a limit cycle that can be perceived in polar coordinates \citep{hafeez2015}.
For our experiments, $u \in [-1, 1]$
and $x_1, x_2 \in [-3, 3]$.

\subsubsection{PINC Analysis}
\label{sec:vdp_analysis}

To find the most suitable configuration for the PINC net to control a dynamical system, we propose first running grid search experiments over hyperparameters, such as the network complexity and the number of data points ($N_t$) and collocation points ($N_f$).

Here, the sampling time is chosen according to the particular dynamics of the Van der Pol oscillator: $T=T_s=0.5s$. At first, we use $N_t=1,000$ and $N_f=100,000$ as they provide a sufficient number of points to train a PINC net.

For training the PINC net, ADAM is used to optimize the loss function for $K_1=500$ epochs, and afterward, L-BFGS is used for $K_2=2,000$ iterations to enhance the stability of the training process. Note that this $K_2$ does not exhaust the training and, as such, it will need to be increased before the final deployment of the PINC net. 
 The parameter $\lambda$ is set empirically so that $\textrm{MSE}_{y}$
and $\textrm{MSE}_{\mathcal{F}}$ are not in disparate scales.
The validation dataset is composed of 1810 points obtained using a randomly generated control action $u$ (e.g., Fig.~\ref{fig:pred_vanderpol}), which is equivalent to $905s$ of simulation, since $T_s=0.5s$.
The validation or generalization error considers the self-loop mode of PINC to compute Eq. \eqref{eq:mse_gen}.

\begin{figure*}[tb!]
    \centering
    \subfloat[Effect of the number of layers and neurons per layer]{
     \includegraphics[width=0.48\linewidth]{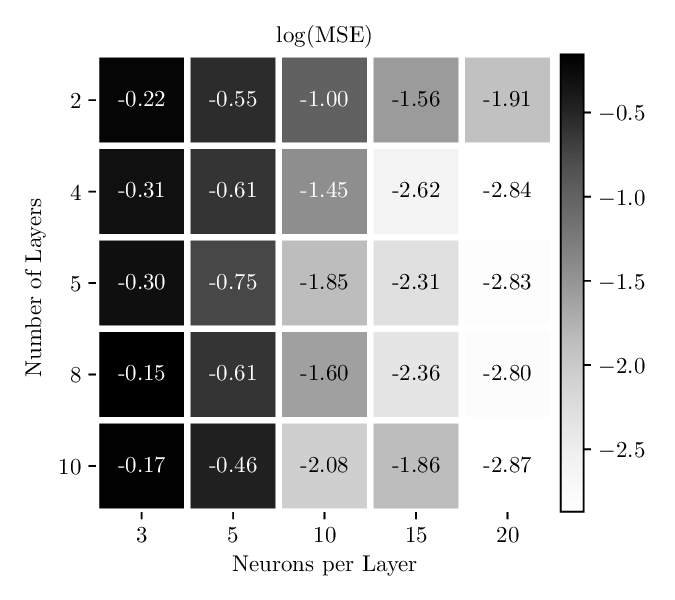}
    \label{fig:vdp_complexity}
     }
    \subfloat[Effect of the number of collocation points $N_f$ and data points $N_t$]{
      \includegraphics[width=0.48\linewidth]{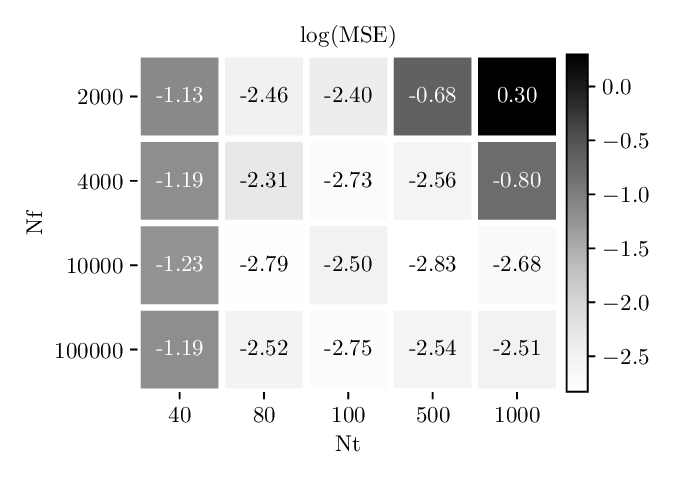}
     \label{fig:vdp_ntnf}
     }
    
    \caption{
    Analysis of the PINC net for the Van der Pol Oscillator. The network training time is fixed to a constant number of iterations.
    The MSE validation error is computed according to Equation (\ref{eq:mse_gen}).
    (a) The $\log_{10}$ of the MSE error as a function of network complexity, averaged over 10 different simulations.
    The best generalization error ($10^{-2.87}$) is achieved with a deep network of 10 layers with 20 neurons each. 
    (b) The effect of the number of collocation points $N_f$ and data points $N_t$ on generalization performance, averaged over 5 different randomly initialized networks.
    }
    \label{fig:exp1_vanderpol}
\end{figure*}

The first experiment analyses the network complexity (Fig.~\ref{fig:vdp_complexity}) and shows the validation MSE using Eq. \eqref{eq:mse_gen} averaged over 10 different random initializations of the network weights.
In general, as the network grows deeper and with more neurons per layer, the performance increases. Besides, layers with 3 or 5 neurons are not sufficient to model the required task. Note that these errors would decrease even further if training had been extended for more epochs (\emph{correcting} the lower performance of the net of 10 layers with 15 neurons each, for instance).
Although the network of 10 layers with 20 neurons each achieves the best performance, we choose to continue the following experiments with a configuration of 4 layers of 20 neurons each, which also showed excellent performance, but with less computational overhead.

In Fig.~\ref{fig:vdp_ntnf}, the proportion between data points and collocation points is investigated. Each error cell in the plot corresponds to the average of 5 different experiments with randomly generated networks.
Clearly, 40 data points are not enough, and the proportion $N_f/N_t$ should be considerably higher than $4$ (hence the dark cells in the upper-right corner of the plot).

\subsubsection{Long-range Simulation}
\label{sec:vdp_simulation}
\eric{this is a new section}

In order to showcase the capacity of long-range simulation of the proposed approach in relation to the conventional PINN, we trained traditional PINNs that have the same complexity of the PINC net, i.e., 4 layers of 20 neurons each,
but that have only one input $t$ as usual for PINNs.
A new PINN is trained for each interval considered $T \in \{0.5, 1, 2, 5, 10\}$ seconds. Note that these PINNs do not allow for arbitrary initial conditions after training, as PINC nets do.
Furthermore, once the interval value $T$ is chosen before training for conventional PINNs, further simulation beyond $T_s$ rapidly deteriorates as we shall see.

The PINNs were trained with the ADAM optimization algorithm for $K_1=500$ epochs initially with a learning rate of $0.0035$, and then other $K_1=700 \times T$ epochs with a learning rate of $0.001$, and finally for $K_2=1,000 \times T$ iterations of the L-FGBS optimization method. In addition, the number of collocation points also increased with the value of $T$, $N_f=5,000 \times T$. Thus, the longer the interval $T$, the longer the training and the higher the number of collocation points employed.
On the other hand, only one PINC network was trained, following the configuration from the previous section, but for longer, as indicated in Fig.~\ref{fig:mse_evolution}.

\begin{figure}[htb!]
    \centering
    \includegraphics[width=1\linewidth]{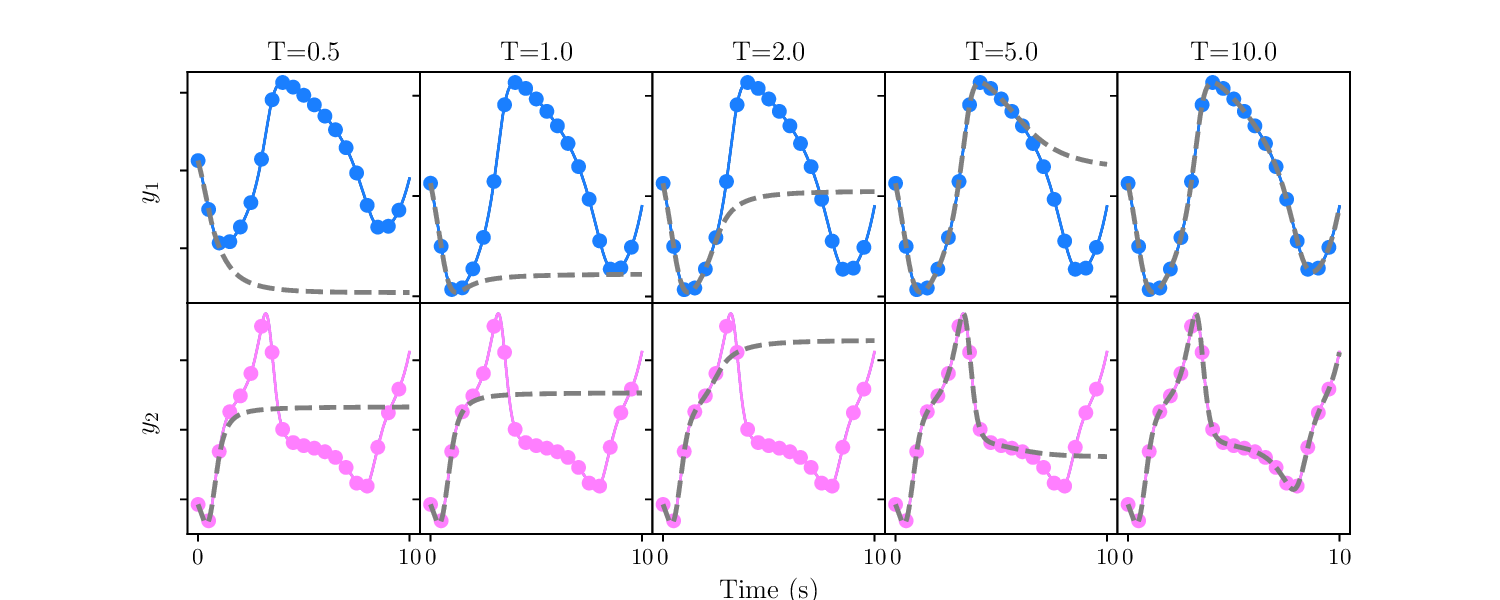}

    \caption{Comparison between conventional PINN (dashed grey line) and proposed PINC (solid blue and pink lines) for long-range simulation of the Van Der Pol oscillator with fixed control input $u$ and fixed initial condition $\mathbf{x}=(x_1,x_2)$ along the simulation.
    The target trajectories for states $x_1$ and $x_2$ are also plot in black solid line, but are completely superimposed by the PINC predictions $y_1$ and $y_2$.
      From left to right, the PINN nets are trained with fixed $T \in \{0.5, 1, 2, 5, 10\}$, \rev{while the PINC net is trained only once with $T=0.5s$ even though it can run for arbitrary longer simulation times not fixed beforehand}. 
    }
    \label{fig:pinn_comparison}
\end{figure}

In Fig.~\ref{fig:pinn_comparison}, the results are shown, which compare the trajectories of the single PINC net that works for any considered interval $T$ (e.g., shorter or longer than $10s$)  with the ones from the PINN networks. The two rows in the plot correspond to the two states of the oscillator. Each subplot involves the training of one new PINN from scratch for a certain $T \in \{0.5, 1, 2, 5, 10\}$, except for the PINC net, which is trained only once.
Besides, each PINN is trained considering a fixed control input $u=0.54$ along the run, with fixed initial conditions $x_1=-2.14$ and  $x_2=0.25$, both randomly chosen. 
Unlike PINC, conventional PINNs would need to be retrained from scratch if a different initial condition or control input is required.

In the plot of Fig.~\ref{fig:pinn_comparison}, the dots in the predicted PINC trajectories, in blue and pink colors, mark the moments at which the final predicted states at $T=0.5s$ are fed back as new initial conditions and input to the network, corresponding to the vertical lines in Fig.~\ref{fig:pincevolution}. 
Although PINC is trained with a fixed $T=0.5s$, its chained (self-loop) prediction can be used to perform long-range simulation \rev{for an arbitrary total simulation time $T$ without fixing it beforehand as with traditional PINNs, whose trajectory is shown by the dashed grey lines in the plots of Fig.~\ref{fig:pinn_comparison}. 
Note that the target true trajectory of the dynamical system, drawn in solid black line, is completely superimposed by the predicted PINC trajectory.
In addition, observe that only the PINN trained specifically with $T=10s$ can simulate without degradation until $10s$, and not beyond that, for the given fixed initial condition and control input.}

\begin{figure}[ht!]
    \centering
    \includegraphics[width=0.43\linewidth]{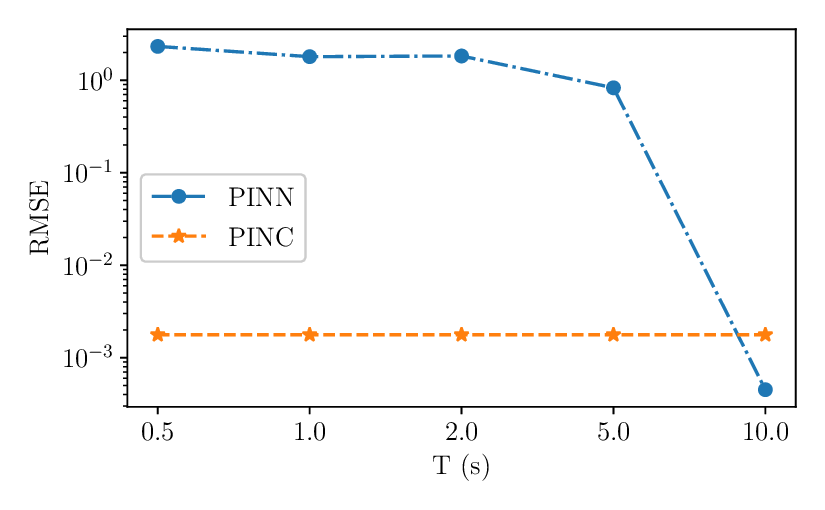}
    \caption{Performance comparison in terms of RMSE between the target trajectory of the Van der Pol oscillator and the predicted trajectory \rev{for the PINN and PINC networks from Fig.~\ref{fig:pinn_comparison} for a simulation of $10s$. The horizontal axis corresponds to the fixed $T$ used for training the PINN network}. See text for more details.
    }
    \label{fig:pinn_comparison_rmse}
\end{figure}

The RMSE error for these experiments are summarized in Fig.~\ref{fig:pinn_comparison_rmse}, making it clear the high prediction error obtained by the conventional PINN in relation to the proposed PINC approach \rev{when the $T$ used for PINN training is lower than $10s$. At $T=10s$, PINN has slightly lower error than PINC, likely because of the small accumulation of prediction errors during self-loop simulation for PINC.}

The control input $u$ was kept fixed here to compare the conventional and the new approach. 
However, PINC can have a variable $u$ along the simulation, yielding an additional advantage for allowing control applications, as showcased in the next section.

\subsubsection{PINC Control}
\label{sec:vdp_control}
\begin{figure}[htb!]
    \centering
    \includegraphics[width=0.6\linewidth]{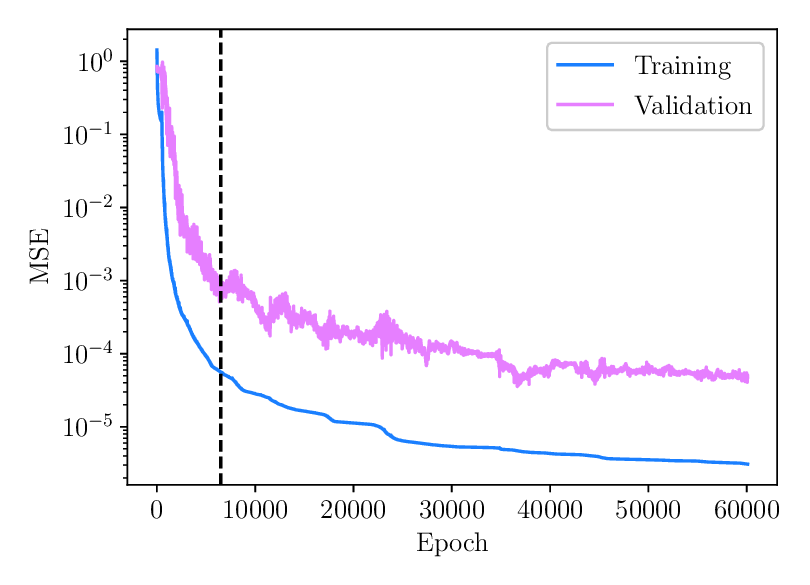}
    \caption{MSE evolution during training of the final PINC net. Previous experiments from Section~\ref{sec:vdp_analysis} stopped training at the vertical dashed line. The validation dataset consists of 1810 points, or $90s$ of simulation since $T=T_s=0.5s$. The validation MSE is more noisy because it is computed on a much smaller dataset and on self-loop mode using Eq. \eqref{eq:mse_gen}.
    }
    \label{fig:mse_evolution}
\end{figure}

The final PINC net is chosen to have 4 hidden layers each of 20 neurons for the Van der Pol oscillator. Besides, we continue setting
$N_t=1,000$, $N_f=100,000$, and $K_1=500$, but the training is extended with
$K_2=20,000$, which allows the MSE to settle in an asymptotic curve (Fig.~\ref{fig:mse_evolution}). For comparison, a vertical black dashed line is plotted in Fig.~\ref{fig:mse_evolution}, indicating the moment at which training would have stopped for earlier experiments from Fig.~\ref{fig:exp1_vanderpol}. Thus, further training allows improving validation error (according to equation \eqref{eq:mse_gen}) at least one order of magnitude. Note that the validation error does not increase permanently as training follows, arguably due to the regularization effect of $\textrm{MSE}_{\mathcal{F}}$ in the loss function.

\begin{figure}[htb!]
    \centering
    \includegraphics[width=0.65\linewidth]{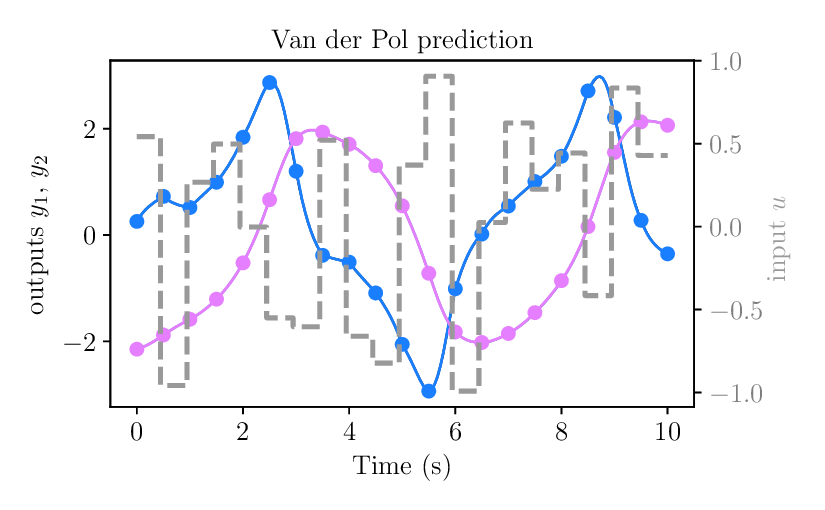}
    \caption{PINC net prediction for the Van der Pol oscillator on test data.
    The grey dashed line gives the randomly generated input $u$, while the predictions for the oscillator states $x_1$ and $x_2$ correspond to the solid blue and pink lines, respectively. The target trajectory, from RK method, is also plot in black, but is not visible as the prediction completely superimposes the former.
    Each dot in the predicted trajectory corresponds to the vertical lines in Fig.~\ref{fig:pincevolution} when the control action and initial state change  (after $T_s=T=0.5s$).
    }
    \label{fig:pred_vanderpol}
\end{figure}

To view the PINC prediction after training, we randomly generate a control input $u$ for $10s$.
In Fig.~\ref{fig:pred_vanderpol}, the predicted trajectory is given for such a control input. With our method, we can directly infer each circle in the trajectory using  \eqref{eq:interface} every $T=0.5s$. The trajectory between two consecutive circles can be predicted by varying the input $t$ of the network and keeping the other inputs $y(0)$ and $u$ fixed. The prediction matches the target trajectory very well as the latter is also plot, but gets superimposed by the former.

\begin{figure}[ht!]
    \centering
    \includegraphics[width=0.53\linewidth]{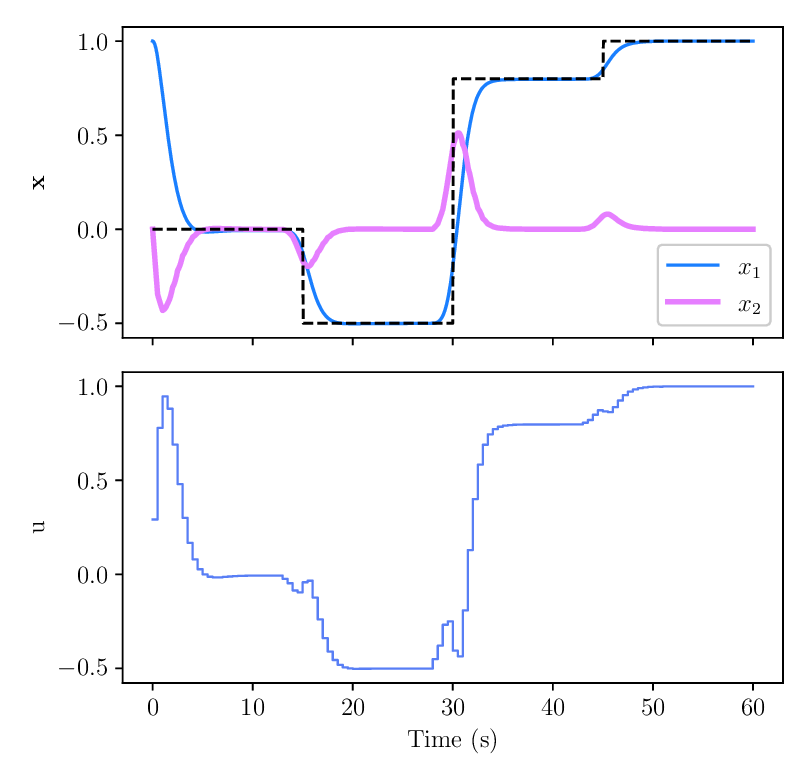}
    \includegraphics[width=0.46\linewidth]{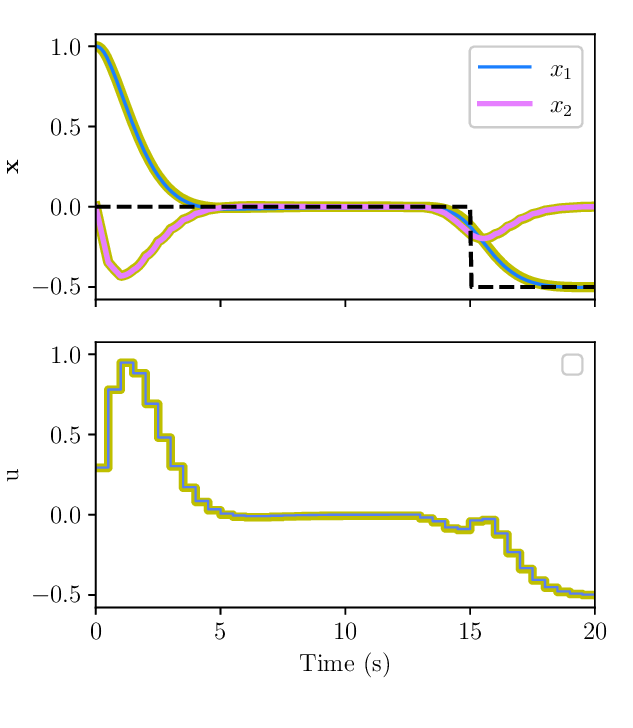}
    \caption{Control of the Van der Pol oscillator with PINC and comparison with the ODE model. The reference trajectory for $x_1$ ($x_2$) is given by a dashed black step signal (fixed at zero), while the controlled variables are the states $x_1$ and $x_2$ given by blue and pink lines.
    The control input $u$ is the manipulated variable in the lower plot, found by MPC.
    Left: simulation totalling 60s exclusively by PINC. 
    Right: comparison with ODE model (plant) as predictive model in yellow thick line during the first 20s.
    See text for more details.}
    \label{fig:control_vanderpol}
\end{figure}

The resulting control from PINC can be seen in Fig.~\ref{fig:control_vanderpol} in a simulation of $60s$, where MPC was employed to find the optimal value of the control input, considering a prediction and control horizon of $5T$ (or $2.5s$).
The control parameters are given as follows: $N_1 = 1$, $N_2 = 5$, $N_u = N_2$, $Q = 10 \mathbf{I}$, and $R = \mathbf{I}$.
Here, the optimization in MPC to find a control input at the current timestep uses the PINC network's predicted trajectory for future timesteps, i.e., for the prediction horizon of $2.5s$. This procedure is repeated for all 120 points of the plotted trajectory.
%
Here, the role of the plant to be controlled is taken by the Van der Pol oscillator, whose states are obtained by an RK integrator.
\eb{
The control performance for a $60s$ simulation is presented in Table~\ref{tab:control}, which also shows the result when the original ODE model is used as the predictive model in NMPC instead of the PINC net. In this case, 
the classic, fourth-order Runge-Kutta method (RK4) is employed
\eaa{as a numerical solution method}
to compute the states of the system for NMPC. 
  This means that practically other approximations to the plant/system are not likely to improve the ODE model itself, thus, justifying our comparison to the baseline NMPC.
Remarkably, PINC achieves practically the same result as the ODE/RK approach in terms of RMSE and IAE, while being slightly faster on average when executed with 10 repetitions on the same desktop computer.
The right plot in Fig.~\ref{fig:control_vanderpol} also shows the NMPC considering the ODE/RK model as the predictive model, in thick yellow line, showing that the lines from PINC and ODE/RK match very well.
}

\subsection{Four Tanks}

\subsubsection{Model}
The four tanks system is a widely used benchmark for multivariate control systems \citep{4tanks_origin}, as it is a nonlinear and multivariate system with some degree of coupling between variables.
By setting its parameters to a given combination of values, it is possible to induce the system to have non-minimum phase transmission zeros, which are an additional difficulty for PID controllers \citep{4tanks_origin}.

\begin{figure}[b!]
	\centering
	\includegraphics[width=0.6\linewidth]{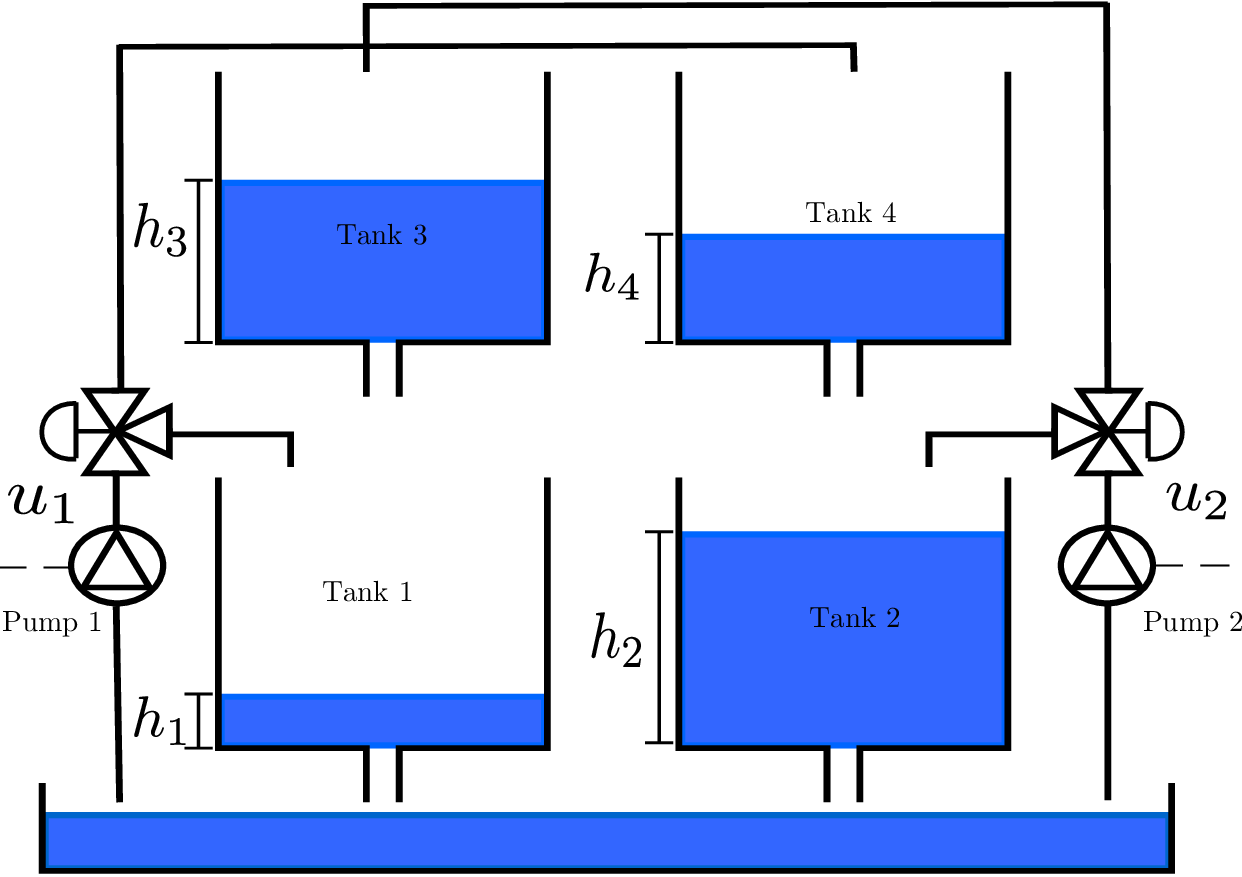}
	\caption{Schematic representation of the four tanks system, from \cite{Brandao2018}.
	}
	\label{fig:four_tanks}
\end{figure}

As Figure \ref{fig:four_tanks} shows,
the four tanks system is composed of four tanks, denoted by the index $i = \{1,2,3,4\}$, and two pumps $j = \{1,2\}$ supplying each tank with water.
Each tank has a cylindric form with a basis area of $A_i$, and an orifice of area $a_i$ at the basis center.
Tank $1$ ($2$) is located right below tank $3$ ($4$), so that the flow $\omega_i$ from the tank above goes directly to the tank below.
Both pumps are linear actuators controlled by the voltage $u_j$ with coefficient $k_j$ converting the voltage into the pump flow.
Pump $1$ ($2$) is associated with a directional valve that distributes the resulting flow into tanks $1$ and $4$ ($2$ and $3$), which is the coupling source in this system.
The directional valves have an opening $\gamma_j$, which is the amount they distribute to the bottom tanks.
The adjustment of $\gamma_j$ is one of the main factors in regulating the control problems associated with the system \citep{4tanks_origin}.
The level of water in each tank is denoted $h_i$.

The following equations govern the four tank system, which are obtained from mass balance:
\begin{subequations}
\begin{align}
    \dot{h}_1 &= \frac{\gamma_1 k_1 u_1 +\omega_3 - \omega_1}{A_1}\\
    \dot{h}_2 &= \frac{\gamma_2 k_2 u_2 +\omega_4 - \omega_2}{A_2}\\
    \dot{h}_3 &= \frac{(1 - \gamma_2) k_2 u_2 -\omega_3}{A_3}\\ 
    \dot{h}_4 &= \frac{(1 - \gamma_1) k_1 u_1 -\omega_4}{A_4}
\end{align}
\end{subequations}
where 
the flow in each tank orifice $\omega_i$ is described by the Bernoulli orifice equation, adding the sole nonlinearity of the system:
\begin{equation}
    \omega_i = a_i\sqrt{2gh_i}
\end{equation}
with $g$ as the acceleration of gravity.
The parameters used for this application are the same as the ones stipulated for the non-minimum phase experiment in \cite{4tanks_origin}.


\subsubsection{PINC Control}

We have performed a similar approach to the first control problem with respect to finding a suitable configuration for network complexity and the proportion between data and collocation points. 
We have observed that $5$ is the minimum number of layers to obtain sufficient prediction performance for the four tanks system, since it is a more complex plant, with multiple inputs and multiple outputs (MIMO) operating at different timescales.
The following experiments consider a PINC net with 5 layers of 20 neurons each.
Besides, we continue setting
$N_t=1,000$, $N_f=100,000$, $K_1=500$, and $K_2=20,000$.
The sampling period is $T=T_s=10s$.  
The control parameters are once again given by $N_1 = 1$, $N_2 = 5$, $N_u = N_2$, $Q = 10 \mathbf{I}$, and $R = \mathbf{I}$.

\begin{figure}[tb!]
    \centering
    \includegraphics[width=0.49\linewidth]{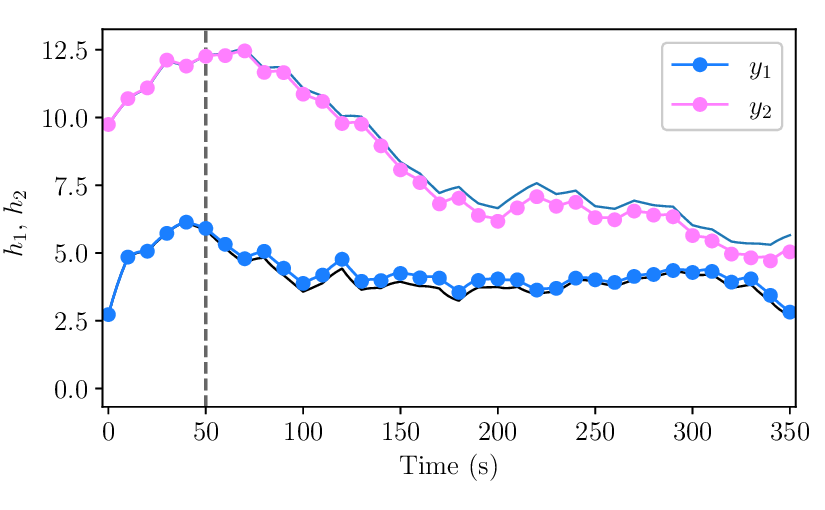}
    \includegraphics[width=0.49\linewidth]{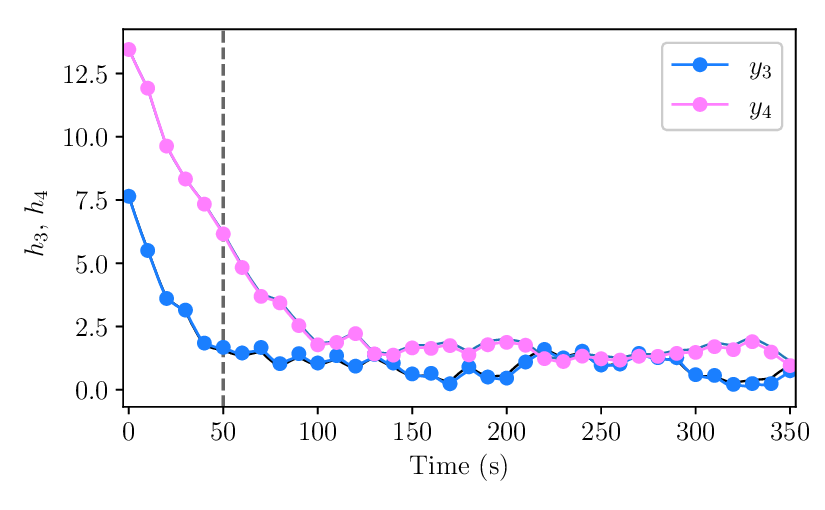}
    \caption{PINC net prediction for the four tanks system on test data, with randomly generated control input signals similar to Fig.~\ref{fig:pred_vanderpol}.
    The predictions for the level of the tanks $h_1$ and $h_2$ correspond to the solid blue and pink lines, respectively. From the RK method, the target trajectories are also plot in dark solid lines without dots.
    At each dot in the predicted trajectory, the PINC net receives new inputs for control action and initial state (every $T_s=T=10s$), as explained in Fig.~\ref{fig:pinc_feedback}. The vertical dashed line indicates the prediction horizon used for MPC.
    }
    \label{fig:pred_fourtanks}
\end{figure}

After training the PINC net, the prediction on test data, with new randomly generated control actions (not shown), is presented in Fig.~\ref{fig:pred_fourtanks}. The deviation in prediction at longer ranges, as seen in the first plot for $h_1$ and $h_2$, is expected, since the network works in self-loop mode, feeding its prediction of the last state back as input for the initial state (Fig.~\ref{fig:pinc_feedback}), every $T=10s$. 
Thus, the error  is accumulated in this chaining procedure. However, MPC uses this trajectory only up to $50s$, equivalent to a prediction horizon of 5 steps, indicated by the vertical dashed line in the figure, and the next optimization procedure in MPC resets the initial state to the true value as obtained by sensors of the real process (Fig.~\ref{fig:pinc_plant}).

\begin{figure}
    \centering
    \includegraphics[width=0.54\linewidth]{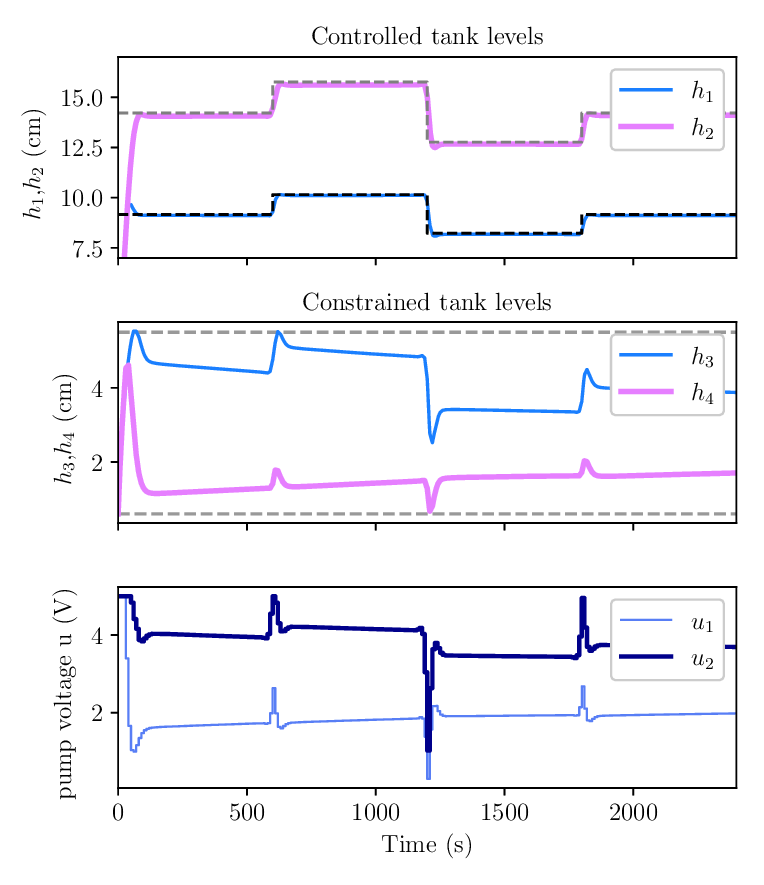}
    \includegraphics[width=0.445\linewidth]{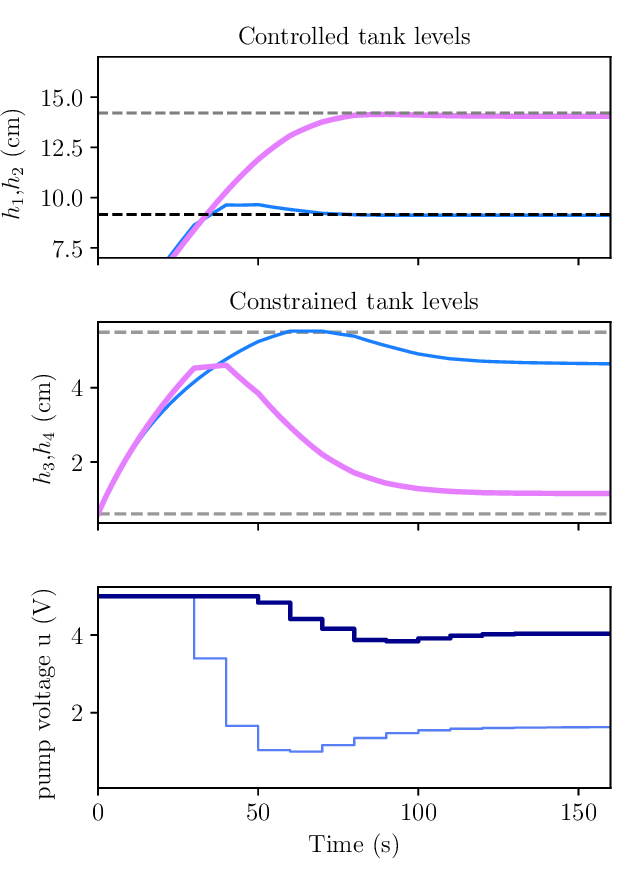}
    \caption{Control of the four tanks system with PINC. 
    The controlled variables are the tank levels $h_1$ and $h_2$ given by blue and pink lines, respectively, whereas
    the reference trajectory for $h_1$ ($h_2$) is given by a dashed black (grey) step signal. 
    The control inputs $\mathbf{u}$ are the manipulated voltages shown in the lower plot, found by MPC.
    Dashed grey horizontal lines represent the lower and upper limits for both $h_3$ and $h_4$.
    Left: simulation totalling $2400$s. Right: close-up on the first $160$s.
    \eaa{The initial conditions for $h_1$ and $h_2$ are $(2, 2)$, which is the minimum of the allowed interval $[2, 20]$.
    See text for more details.}
    }
    \label{fig:control_fourtanks}
\end{figure}

\begin{figure}
    \centering
    \includegraphics[width=0.8\linewidth]{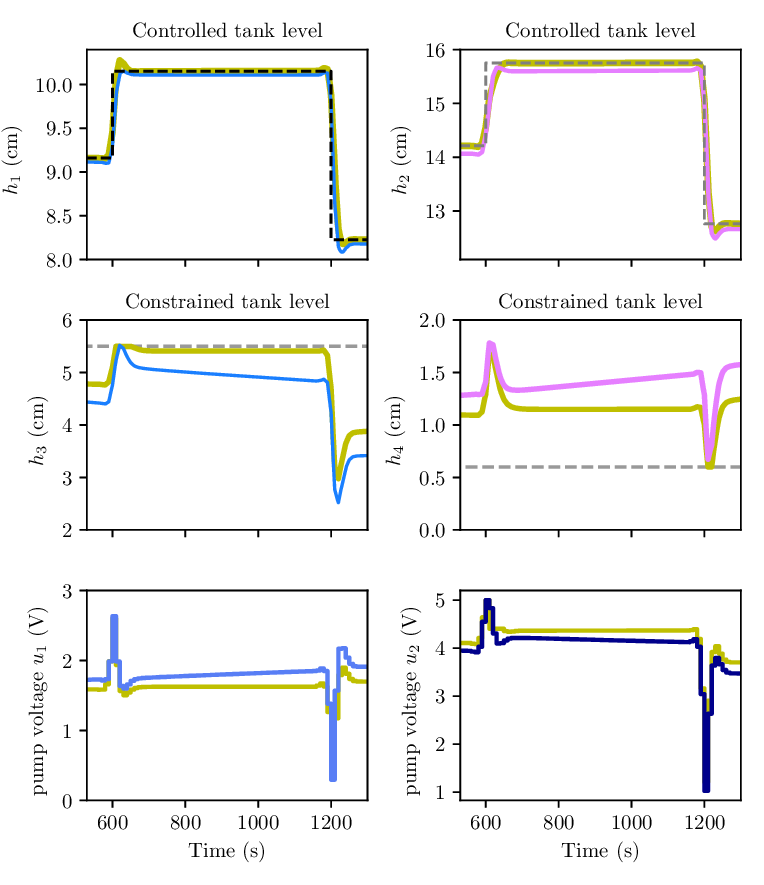}
    \caption{Control of the four tanks system with PINC and the RK/ODE model for timesteps 500s to 1300s from Fig.~\ref{fig:control_fourtanks}. 
    All the yellow lines correspond to the simulation where the ODE model itself is used as predictive model (using a numerical solution at each timestep in the prediction horizon), whereas the remaining lines belong to the simulation with PINC net as predictive model.
    }
    \label{fig:control_fourtanks_comparison}
\end{figure}

PINC's control employs prediction and control horizons both of 5 steps ($50s$ in simulation time) for this four tanks system. Besides, both $h_3$ and $h_4$ tank levels are constrained to the interval $[0.6, 5.5]$cm. 
The results are shown in Fig.~\ref{fig:control_fourtanks}, where the controlled and constrained tank levels are presented in the first two topmost plots, and the control action found by MPC is shown at the bottom plot. The plots on the right-hand side show a close-up during the initial $160$s of the simulation.
The control was successful in spite of the constrains imposed on $h_3$ and $h_4$ (which were respected) and some minor error in steady regime, which can be corrected by adding the calculation of a correction factor through filtering the error between the measurement and the network prediction, as done in \cite{Jordanou2018} for a recurrent network.
\eb{In Fig.~\ref{fig:control_fourtanks_comparison}, we use the same simulation setup, focusing on the timesteps between 500s and 1300s, to compare with the response (in yellow color) of the control using the plant reference model as predictive model in MPC.
  This ODE/RK-based model is the reference model that represent the plant itself, which justifies the negligible steady-state regime error observed in the figure. 
We can notice that the PINC simulation is very close to the nominal MPC given by the ODE/RK model. This comparison suffices as another NMPC would employ an approximation of the ODE/RK model as a predictive model.
}

The control performance in terms of RMSE and IAE is shown in Table~\ref{tab:control}. Although IAE seems to differ more between PINC and ODE/RK, RMSE errors for both methods are almost equivalent. 
The average time spent on a desktop computer for the complete control simulation using PINC, repeated 10 times, $10.85s$, is $23.3\%$ inferior than  using the ODE of the four tanks as a model for MPC ($14.15s$), which is remarkable given the architecture of 5 hidden layers with 20 neurons each that is used for PINC.

\begin{table}[tb]
    \centering
        \caption{Results for the control experiments}    \label{tab:control}
    \begin{tabular}{c|ccc|ccc}
\multicolumn{1}{c}{} & \multicolumn{3}{c|}{Van der Pol} & \multicolumn{3}{c}{Four tanks} \\
\multicolumn{1}{c}{} & \multicolumn{3}{c|}{(4 layers of 20 units)} 
                     & \multicolumn{3}{c}{(5 layers of 20 units)} \\ \cline{2-4} \cline{5-7}
 \multicolumn{1}{c}{}  & RMSE & IAE & time ($s$)       & RMSE  & IAE & time ($s$)    \\ \hline         
    PINC       & 0.15 & 123.6 & $\mathbf{3.32 \pm 0.15} $   & 0.811 & 876 & $\mathbf{10.85 \pm 0.14}$ \\
    \hline
    ODE / RK   & 0.15 & \textbf{122} & $3.41 \pm 0.04 $     & \textbf{0.807} & \textbf{544} & $14.15 \pm 0.13$
    \end{tabular}
\end{table}

\subsubsection{Sensitivity to Perturbations}
 
The PINC approach, as it was introduced in this work, does not have an inherent method to deal with modeling errors and completely reject disturbances, making these points valid for future research. 
  Nonetheless, some of robustness to model mismatch and disturbances can be implemented in the control algorithm by means of error correction filtering  \citep{Camacho2013} and Kalman filters \citep{Brown:Book:KF:1992}.
 
While works such as \cite{Cheng2021} and \cite{Wei2021} are focused on proving stability theoretically
through the use Lyapunov functions, 
we showcase the PINC robustness experimentally since our focus is more application oriented.
In order to test the robustness of the proposed formulation to parameter mismatch, a sensitivity analysis was performed for the four tanks scenarios previously presented.
 The analysis was performed assuming random deviations in the values of the $k_1$ and $k_2$ parameters. 
The perturbed values of $\widetilde{k}_1$ and $\widetilde{k}_2$ are sampled from uniform probability distributions $U_{5\%}(a,b) = [0.95 x, 1.05 x]$, with $x$ being the nominal value in which the PINN was trained.

\begin{figure}
    \centering
    \subfloat[\textbf{5 layers} of 20 neurons each (left: $k_1$ and $k_2$ perturbed; right: initial condition perturbed)
    ]{
    \includegraphics[width=0.4\linewidth]{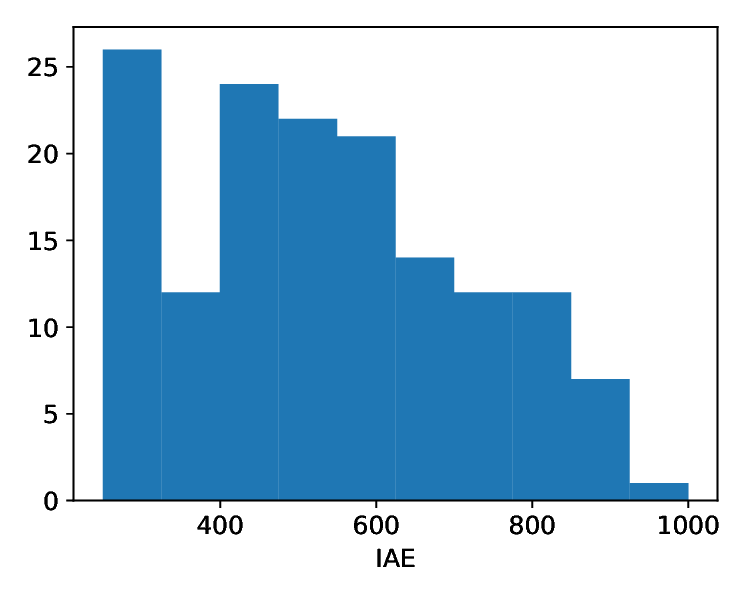}
    \includegraphics[width=0.4\linewidth]{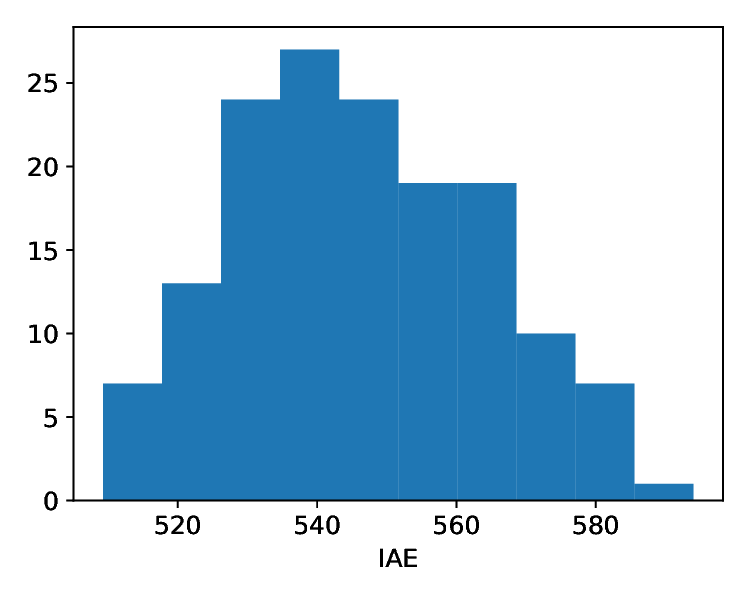}
    }
    \\
    \subfloat[\textbf{8 layers} of 20 neurons each
    (left: $k_1$ and $k_2$ perturbed; right: initial condition perturbed)
    ]{
    \includegraphics[width=0.4\linewidth]{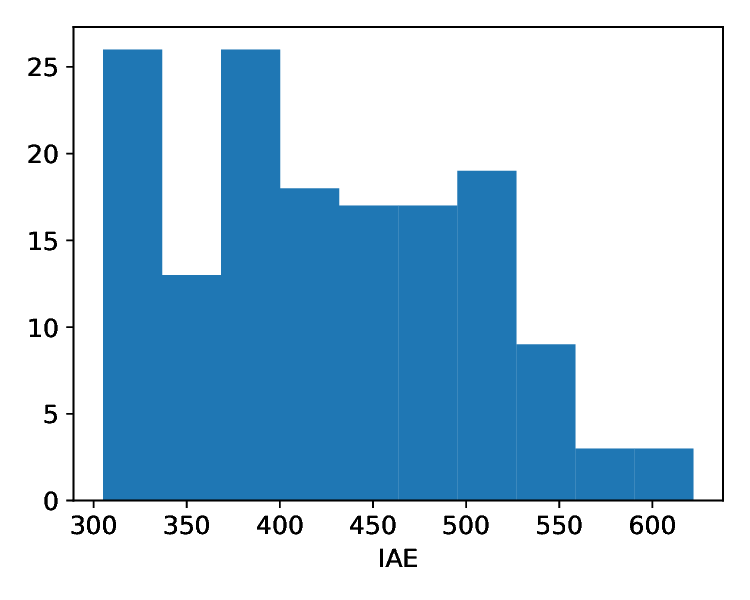}
    \includegraphics[width=0.4\linewidth]{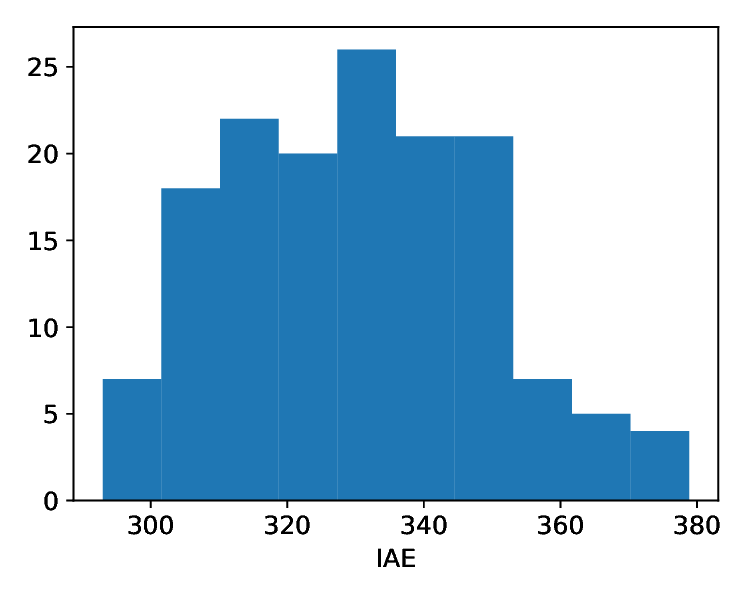}
    }
    
    \caption{Sensitivity to modeling errors (left) and to initial conditions (right) for the MPC of the four tanks system with PINC net as the model. 
    For each of the plots, 151 runs of the MPC algorithm using a trained PINC net of 5 layers (top plots) and 8 layers (bottom plots) are executed. The resulting IAEs between the references (as in Fig.~\ref{fig:control_fourtanks}) and the controlled signals $h_1$ and $h_2$ are computed and shown in a histogram. The initial conditions are $h_1=h_2=9$ , which is the middle point of the allowed interval.
    %
    %
    %
    }
    \label{fig:ft_sens_modelerror}
\end{figure}

\begin{figure}
  \centering
    \includegraphics[width=0.5\linewidth]{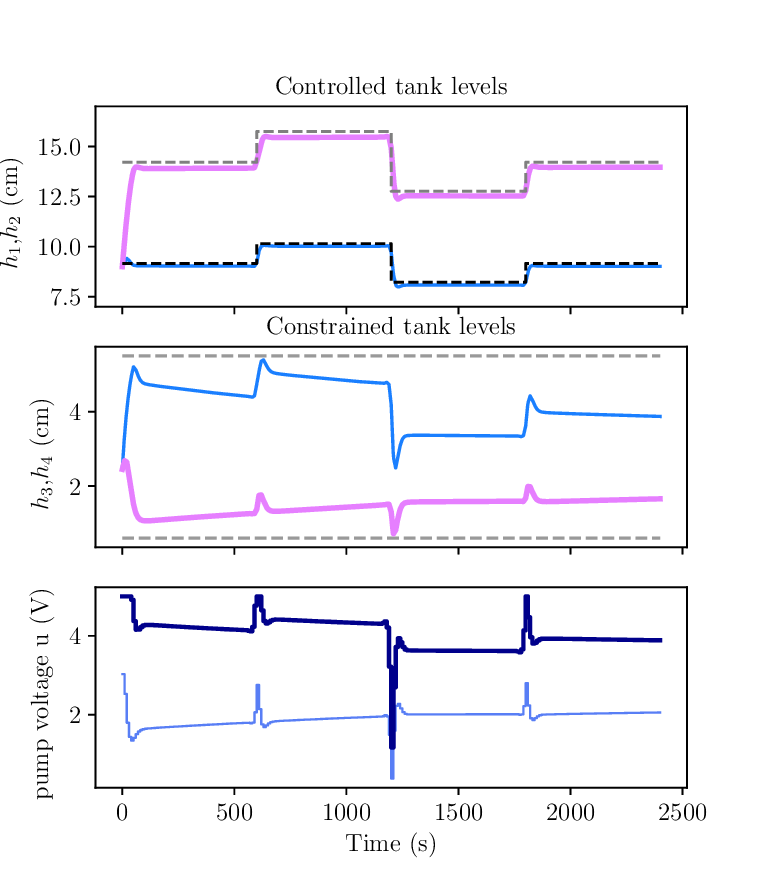}
    \caption{Control of the four tanks system with PINC as in Fig.~\ref{fig:control_fourtanks}, but with maximum perturbation of 5\% in $k_1$ and $k_2$ parameters of the model being controlled. Initial conditions as in Fig.~\ref{fig:ft_sens_modelerror}, that is, $h_1=h_2=9$.
    Control results were adequate even though the IAE was 1022, which is in the tail of the histogram from Fig.~\ref{fig:ft_sens_modelerror}(a) left plot.
    }
    \label{fig:control_perturb}
\end{figure}

Altogether, 151 simulations were carried out, injecting the deviations in the system. The results are shown in first column of Fig. \ref{fig:ft_sens_modelerror} for two different networks, one with 5 hidden layers of 20 neurons each, and another one with 8 hidden layers of 20 neurons each. 
Despite the random variation of the parameters $k_1$ and $k_2$, it can be seen that the IAE of the system has variations within a tolerance range considered adequate.
Fig.~\ref{fig:control_perturb} shows the control of the four tanks when the plant controlled has the maximum deviation of 5\% in $k_1$ and $k_2$ parameters, showcasing that the perturbation only slightly bias the trajectories.

\eaa{A second experiment consisted of perturbing the initial condition
with a uniform distribution. The perturbed initial conditions for $h_1$ and $h_2$ are sampled from
$U_{5\%}(a,b) = [0.95 x, 1.05 x]$, with $x$ being the nominal value $9$ for both states. The results are shown in the second column of Fig.~\ref{fig:ft_sens_modelerror}. The peak of the histogram approximately coincides with the IAE obtained by the unperturbed model of the plant. Thus, other initial conditions can imply relatively lower or higher IAE.
Notice that the bottom plots show instances with lower IAE, evidencing the higher accuracy of a deeper network, with 8 hidden layers, in this particular situation.
}

In summary, the sensitivity experiments imply that the system does not lose much performance in terms of IAE.
In fact, the occurrence of lower IAE simulations is actually higher.
Since the PINC control strategy has no integrators, a small steady state error that depends on model match is expected.
As there is more model mismatch when the parameters $k_1$ and $k_2$ distance themselves from the nominal ones, the steady-state error is expected to be higher but still within an acceptable range of IAEs.

\subsection{Discussion}
\eaa{
In the context of ODE simulation, our proposal, after the network is trained, has shown that it is possible to speed up the runtime of these simulations (up to 30\% on average).
 With further enhancements in the network inference (e.g., parallelization) or in the case of extending our method to PDEs, we envision an even higher gain (e.g., 10x faster as reported in the literature for simulation of PDEs with PINNs).}

\eaa{
One of the main obstacles to having a fully effective simulation from a PINC network is the long training of such networks. Nonetheless, this is a common issue in most of the proposals dealing with deep learning and specially with PINNs. However, after training, the network can predict directly any state in the range $[0,T]$ without requiring an integration with intermediate points as in numerical simulation methods.
}

\eaa{
We noticed that a precise optimization algorithm towards the end of the training (e.g., L-BFGS) is essential in obtaining a precise model.
Besides, preliminary work in identifying more complex plants (e.g., in oil and gas industry) show that skip connections \citep{Lee2015,He2016} can help the training of deep networks by helping to backpropagate the gradient to the deepest layers during training, further improving the precision of the final trained model.
}

\eaa{
Furthermore, challenges to the learning of PINNs can arise from discontinuities in the ODE equations that model the plant, such as the presence of the $\max$ operator. Also, the random initialization of the weights of neural networks may cause different results and also render invalid arguments to functions such as the square root, if present in the ODE equations. Notice that some fixes or workarounds can be applied in these cases.
}

\section{Conclusion}
\label{sec:conclusion}


We have proposed a new framework that makes Physics-Informed Neural Networks (PINNs) amenable to control methods, such as MPC, opening a wide range of application opportunities.
This Physics-Informed Neural Nets-based Control (PINC) approach allows a PINN to work for longer-range time intervals that are not fixed beforehand, without severe prediction degradation as it normally does, and makes it easy to employ such networks in MPC applications.
In control applications,  this framework (a) provides a way to identify a system by integrating collected data from a plant with a priori expert knowledge in the form of ordinary differential equations;
\eaa{(b) can simulate differential equations faster than numerical solution methods, specially if extended to Partial Differential Equations (PDE), making PINNs more appealing to real-time control applications.}
Although we only used initial conditions as real training data, we foresee that using additional sparse data will make the training of PINC nets much faster.

In future work, we intend to extend the framework to systems described by Differential-Algebraic Equations (DAEs) and PDEs, and systems for which prior knowledge is uncertain (unknown parameters)
as well as apply PINC to industrial control problems, such as in the oil and gas industry, for which some prior knowledge of ODEs are known in addition to very noisy \eaa{or sparse} data. 
\eaa{We expect that the reduction in the computational burden in using PINC for control scenarios
will be even more relevant in comparison to the numerical solution approach, as the model becomes increasingly more complex or in the case of models described by PDEs.} 
\eaa{Finally, we envision that the application of system identification in an industrial setting will expand if we use complementary sources of information for training deep networks, that is, by using physical laws and historical sparse data}, 
making feasible a wide range of previously unsolved applications in systems and control.

\section*{Acknowledgments}

This work was funded in part by CNPq (Grant 308624/2021-1)  and FAPESC (Grant 2021TR2265).

\bibliography{mybib.bib}

\begin{thebibliography}{49}
\expandafter\ifx\csname natexlab\endcsname\relax\def\natexlab#1{#1}\fi
\providecommand{\url}[1]{\texttt{#1}}
\providecommand{\href}[2]{#2}
\providecommand{\path}[1]{#1}
\providecommand{\DOIprefix}{doi:}
\providecommand{\ArXivprefix}{arXiv:}
\providecommand{\URLprefix}{URL: }
\providecommand{\Pubmedprefix}{pmid:}
\providecommand{\doi}[1]{\href{http://dx.doi.org/#1}{\path{#1}}}
\providecommand{\Pubmed}[1]{\href{pmid:#1}{\path{#1}}}
\providecommand{\bibinfo}[2]{#2}
\ifx\xfnm\relax \def\xfnm[#1]{\unskip,\space#1}\fi
\bibitem[{{\AA}kesson \& Toivonen(2006)}]{Aakesson2006}
\bibinfo{author}{{\AA}kesson, B.~M.}, \& \bibinfo{author}{Toivonen, H.~T.}
  (\bibinfo{year}{2006}).
\newblock \bibinfo{title}{A neural network model predictive controller}.
\newblock {\it \bibinfo{journal}{Journal of Process Control}\/},  {\it
  \bibinfo{volume}{16}\/}, \bibinfo{pages}{937--946}.
\bibitem[{Andersson et~al.(2012)Andersson, Åkesson \& Diehl}]{andersson2012}
\bibinfo{author}{Andersson, J.}, \bibinfo{author}{Åkesson, J.}, \&
  \bibinfo{author}{Diehl, M.} (\bibinfo{year}{2012}).
\newblock \bibinfo{title}{Dynamic optimization with {CasADi}}.
\newblock In {\it \bibinfo{booktitle}{Proceedings of the IEEE Conference on
  Decision and Control}\/} (pp. \bibinfo{pages}{681--686}).
\newblock \DOIprefix\doi{10.1109/CDC.2012.6426534}.
\bibitem[{Andrew \& Gao(2007)}]{Andrew2007}
\bibinfo{author}{Andrew, G.}, \& \bibinfo{author}{Gao, J.}
  (\bibinfo{year}{2007}).
\newblock \bibinfo{title}{Scalable training of {L1}-regularized log-linear
  models}.
\newblock In {\it \bibinfo{booktitle}{Proceedings of the 24th International
  Conference on Machine Learning}\/} ICML'07 (pp. \bibinfo{pages}{33--40}).
\newblock \bibinfo{address}{New York, NY, USA}: \bibinfo{publisher}{Association
  for Computing Machinery}.
\newblock \DOIprefix\doi{10.1145/1273496.1273501}.
\bibitem[{Antonelo et~al.(2021)Antonelo, Camponogara, Seman, de~Souza, Jordanou
  \& Hubner}]{Antonelo2021}
\bibinfo{author}{Antonelo, E.~A.}, \bibinfo{author}{Camponogara, E.},
  \bibinfo{author}{Seman, L.~O.}, \bibinfo{author}{de~Souza, E.~R.},
  \bibinfo{author}{Jordanou, J.~P.}, \& \bibinfo{author}{Hubner, J.~F.}
  (\bibinfo{year}{2021}).
\newblock \bibinfo{title}{Physics-informed neural nets for control of dynamical
  systems}.
\newblock \URLprefix \url{https://arxiv.org/abs/2104.02556}.
  \DOIprefix\doi{10.48550/ARXIV.2104.02556}.
\bibitem[{Biegler(2010)}]{Biegler:SIAM:2010}
\bibinfo{author}{Biegler, L.~T.} (\bibinfo{year}{2010}).
\newblock {\it \bibinfo{title}{Nonlinear Programming: Concepts, Algorithms, and
  Applications to Chemical Processes}\/}.
\newblock \bibinfo{address}{Philadelphia}: \bibinfo{publisher}{SIAM}.
\bibitem[{Bishop(2006)}]{Bishop2006}
\bibinfo{author}{Bishop, C.~M.} (\bibinfo{year}{2006}).
\newblock {\it \bibinfo{title}{Pattern Recognition and Machine Learning
  (Information Science and Statistics)}\/}.
\newblock \bibinfo{publisher}{Springer}.
\bibitem[{Brand{\~{a}}o(2018)}]{Brandao2018}
\bibinfo{author}{Brand{\~{a}}o, A. S.~M.} (\bibinfo{year}{2018}).
\newblock {\it \bibinfo{title}{Controle Preditivo com Gera{\c{c}}{\~{a}}o de
  C{\'{o}}digo: Um Estudo Comparativo}\/}.
\newblock Master's thesis Universidade Federal de Santa Catarina.
\bibitem[{Brown \& Hwang(1992)}]{Brown:Book:KF:1992}
\bibinfo{author}{Brown, R.~G.}, \& \bibinfo{author}{Hwang, P. Y.~C.}
  (\bibinfo{year}{1992}).
\newblock {\it \bibinfo{title}{Introduction to Random Signals and Applied
  Kalman Filtering}\/}.
\newblock \bibinfo{publisher}{John Wiley \& Sons}.
\bibitem[{Camacho \& Bordons(2013)}]{Camacho2013}
\bibinfo{author}{Camacho, E.~F.}, \& \bibinfo{author}{Bordons, C.}
  (\bibinfo{year}{2013}).
\newblock {\it \bibinfo{title}{Model Predictive Control}\/}.
\newblock \bibinfo{publisher}{Springer Science \& Business Media}.
\bibitem[{Cavagnari et~al.(1999)Cavagnari, Magni \& Scattolini}]{Cavagnari1999}
\bibinfo{author}{Cavagnari, L.}, \bibinfo{author}{Magni, L.}, \&
  \bibinfo{author}{Scattolini, R.} (\bibinfo{year}{1999}).
\newblock \bibinfo{title}{Neural network implementation of nonlinear
  receding-horizon control}.
\newblock {\it \bibinfo{journal}{Neural computing \& applications}\/},  {\it
  \bibinfo{volume}{8}\/}, \bibinfo{pages}{86--92}.
\bibitem[{Chen \& Liu(2021)}]{Chen2021}
\bibinfo{author}{Chen, J.}, \& \bibinfo{author}{Liu, Y.}
  (\bibinfo{year}{2021}).
\newblock \bibinfo{title}{Probabilistic physics-guided machine learning for
  fatigue data analysis}.
\newblock {\it \bibinfo{journal}{Expert Systems with Applications}\/},  {\it
  \bibinfo{volume}{168}\/}, \bibinfo{pages}{114316}.
  \DOIprefix\doi{10.1016/j.eswa.2020.114316}.
\bibitem[{Cheng et~al.(2021)Cheng, Chen, Stojanovic \& He}]{Cheng2021}
\bibinfo{author}{Cheng, P.}, \bibinfo{author}{Chen, M.},
  \bibinfo{author}{Stojanovic, V.}, \& \bibinfo{author}{He, S.}
  (\bibinfo{year}{2021}).
\newblock \bibinfo{title}{Asynchronous fault detection filtering for piecewise
  homogenous markov jump linear systems via a dual hidden markov model}.
\newblock {\it \bibinfo{journal}{Mechanical Systems and Signal Processing}\/},
  {\it \bibinfo{volume}{151}\/}, \bibinfo{pages}{107353}. \URLprefix
  \url{https://doi.org/10.1016/j.ymssp.2020.107353}.
  \DOIprefix\doi{10.1016/j.ymssp.2020.107353}.
\bibitem[{Eren et~al.(2017)Eren, Prach, Koçer, Raković, Kayacan \&
  Açıkmeşe}]{aerospace_mpc}
\bibinfo{author}{Eren, U.}, \bibinfo{author}{Prach, A.},
  \bibinfo{author}{Koçer, B.~B.}, \bibinfo{author}{Raković, S.~V.},
  \bibinfo{author}{Kayacan, E.}, \& \bibinfo{author}{Açıkmeşe, B.}
  (\bibinfo{year}{2017}).
\newblock \bibinfo{title}{Model predictive control in aerospace systems:
  Current state and opportunities}.
\newblock {\it \bibinfo{journal}{Journal of Guidance, Control, and
  Dynamics}\/},  {\it \bibinfo{volume}{40}\/}, \bibinfo{pages}{1541--1566}.
  \DOIprefix\doi{10.2514/1.G002507}.
\bibitem[{Gill et~al.(2005)Gill, Murray \& Saunders}]{Gill:SNOPT:2005}
\bibinfo{author}{Gill, P.~E.}, \bibinfo{author}{Murray, W.}, \&
  \bibinfo{author}{Saunders, M.~A.} (\bibinfo{year}{2005}).
\newblock \bibinfo{title}{{SNOPT}: An {SQP} algorithm for large-scale
  constrained optimization}.
\newblock {\it \bibinfo{journal}{SIAM Review}\/},  {\it
  \bibinfo{volume}{47}\/}, \bibinfo{pages}{99--131}.
  \DOIprefix\doi{10.1137/S0036144504446096}.
\bibitem[{Gokhale et~al.(2022)Gokhale, Claessens \& Develder}]{Gokhale2022}
\bibinfo{author}{Gokhale, G.}, \bibinfo{author}{Claessens, B.}, \&
  \bibinfo{author}{Develder, C.} (\bibinfo{year}{2022}).
\newblock \bibinfo{title}{Physics informed neural networks for control oriented
  thermal modeling of buildings}.
\newblock {\it \bibinfo{journal}{Applied Energy}\/},  {\it
  \bibinfo{volume}{314}\/}, \bibinfo{pages}{118852}.
\bibitem[{Gr\"une \& Pannek(2011)}]{Grune:NMPC:2011}
\bibinfo{author}{Gr\"une, L.}, \& \bibinfo{author}{Pannek, J.}
  (\bibinfo{year}{2011}).
\newblock {\it \bibinfo{title}{Nonlinear Model Predictive Control: Theory and
  Algorithms}\/}.
\newblock \bibinfo{publisher}{Springer}.
\bibitem[{Harp et~al.(2021)Harp, O’Malley, Yan \& Pawar}]{Harp2021}
\bibinfo{author}{Harp, D.~R.}, \bibinfo{author}{O’Malley, D.},
  \bibinfo{author}{Yan, B.}, \& \bibinfo{author}{Pawar, R.}
  (\bibinfo{year}{2021}).
\newblock \bibinfo{title}{On the feasibility of using physics-informed machine
  learning for underground reservoir pressure management}.
\newblock {\it \bibinfo{journal}{Expert Systems with Applications}\/},  {\it
  \bibinfo{volume}{178}\/}, \bibinfo{pages}{115006}.
  \DOIprefix\doi{10.1016/j.eswa.2021.115006}.
\bibitem[{He et~al.(2016)He, Zhang, Ren \& Sun}]{He2016}
\bibinfo{author}{He, K.}, \bibinfo{author}{Zhang, X.}, \bibinfo{author}{Ren,
  S.}, \& \bibinfo{author}{Sun, J.} (\bibinfo{year}{2016}).
\newblock \bibinfo{title}{Deep residual learning for image recognition}.
\newblock In {\it \bibinfo{booktitle}{Proceedings of the IEEE Conference on
  Computer Vision and Pattern Recognition}\/} (pp. \bibinfo{pages}{770--778}).
\bibitem[{Hertneck et~al.(2018)Hertneck, K{\"o}hler, Trimpe \&
  Allg{\"o}wer}]{Hertneck2018}
\bibinfo{author}{Hertneck, M.}, \bibinfo{author}{K{\"o}hler, J.},
  \bibinfo{author}{Trimpe, S.}, \& \bibinfo{author}{Allg{\"o}wer, F.}
  (\bibinfo{year}{2018}).
\newblock \bibinfo{title}{Learning an approximate model predictive controller
  with guarantees}.
\newblock {\it \bibinfo{journal}{IEEE Control Systems Letters}\/},  {\it
  \bibinfo{volume}{2}\/}, \bibinfo{pages}{543--548}.
\bibitem[{Iserles(1996)}]{Iserles:Book:1996}
\bibinfo{author}{Iserles, A.} (\bibinfo{year}{1996}).
\newblock {\it \bibinfo{title}{A First Course in the Numerical Analysis of
  Differential Equations}\/}.
\newblock \bibinfo{publisher}{Cambridge University Press}.
\bibitem[{Johansson(2000)}]{4tanks_origin}
\bibinfo{author}{Johansson, K.} (\bibinfo{year}{2000}).
\newblock \bibinfo{title}{The quadruple-tank process: A multivariable
  laboratory process with an adjustable zero}.
\newblock {\it \bibinfo{journal}{IEEE Transactions on Control Systems
  Technology}\/},  {\it \bibinfo{volume}{8}\/}, \bibinfo{pages}{456--465}.
  \DOIprefix\doi{10.1109/87.845876}.
\bibitem[{Jordanou et~al.(2021)Jordanou, Antonelo \&
  Camponogara}]{Jordanou2021}
\bibinfo{author}{Jordanou, J.~P.}, \bibinfo{author}{Antonelo, E.~A.}, \&
  \bibinfo{author}{Camponogara, E.} (\bibinfo{year}{2021}).
\newblock \bibinfo{title}{Echo state networks for practical nonlinear model
  predictive control of unknown dynamic systems}.
\newblock {\it \bibinfo{journal}{IEEE Transactions on Neural Networks and
  Learning Systems}\/},  (pp. \bibinfo{pages}{1--15}).
  \DOIprefix\doi{10.1109/TNNLS.2021.3136357}.
\bibitem[{Jordanou et~al.(2018)Jordanou, Camponogara, Antonelo \&
  Aguiar}]{Jordanou2018}
\bibinfo{author}{Jordanou, J.~P.}, \bibinfo{author}{Camponogara, E.},
  \bibinfo{author}{Antonelo, E.~A.}, \& \bibinfo{author}{Aguiar, M. A.~S.}
  (\bibinfo{year}{2018}).
\newblock \bibinfo{title}{Nonlinear model predictive control of an oil well
  with echo state networks}.
\newblock {\it \bibinfo{journal}{IFAC-PapersOnLine}\/},  {\it
  \bibinfo{volume}{51}\/}, \bibinfo{pages}{13--18}.
  \DOIprefix\doi{10.1016/j.ifacol.2018.06.348}.
\bibitem[{Kingma \& Ba(2014)}]{Kingma2014}
\bibinfo{author}{Kingma, D.~P.}, \& \bibinfo{author}{Ba, J.}
  (\bibinfo{year}{2014}).
\newblock \bibinfo{title}{{ADAM}: A method for stochastic optimization}.
\newblock {\it \bibinfo{journal}{arXiv preprint arXiv:1412.6980}\/}, .
\bibitem[{Kumar et~al.(2021)Kumar, Ridha, Narahari \& Ilyas}]{Kumar2021}
\bibinfo{author}{Kumar, A.}, \bibinfo{author}{Ridha, S.},
  \bibinfo{author}{Narahari, M.}, \& \bibinfo{author}{Ilyas, S.~U.}
  (\bibinfo{year}{2021}).
\newblock \bibinfo{title}{Physics-guided deep neural network to characterize
  non-newtonian fluid flow for optimal use of energy resources}.
\newblock {\it \bibinfo{journal}{Expert Systems with Applications}\/},  (p.
  \bibinfo{pages}{115409}). \DOIprefix\doi{10.1016/j.eswa.2021.115409}.
\bibitem[{{\L}awry{\'{n}}czuk(2014)}]{Maciej2014}
\bibinfo{author}{{\L}awry{\'{n}}czuk, M.} (\bibinfo{year}{2014}).
\newblock {\it \bibinfo{title}{Computationally Efficient Model Predictive
  Control Algorithms}\/}.
\newblock \bibinfo{publisher}{Springer International Publishing}.
\bibitem[{Lee et~al.(2015)Lee, Xie, Gallagher, Zhang \& Tu}]{Lee2015}
\bibinfo{author}{Lee, C.-Y.}, \bibinfo{author}{Xie, S.},
  \bibinfo{author}{Gallagher, P.}, \bibinfo{author}{Zhang, Z.}, \&
  \bibinfo{author}{Tu, Z.} (\bibinfo{year}{2015}).
\newblock \bibinfo{title}{Deeply-supervised nets}.
\newblock In {\it \bibinfo{booktitle}{Artificial Intelligence and
  Statistics}\/} (pp. \bibinfo{pages}{562--570}).
\newblock \bibinfo{organization}{PMLR}.
\bibitem[{Liu \& Wang(2021)}]{Liu2021}
\bibinfo{author}{Liu, X.-Y.}, \& \bibinfo{author}{Wang, J.-X.}
  (\bibinfo{year}{2021}).
\newblock \bibinfo{title}{Physics-informed dyna-style model-based deep
  reinforcement learning for dynamic control}.
\newblock {\it \bibinfo{journal}{Proceedings of the Royal Society A}\/},  {\it
  \bibinfo{volume}{477}\/}, \bibinfo{pages}{20210618}.
\bibitem[{Meng et~al.(2020)Meng, Li, Zhang \& Karniadakis}]{Meng2020}
\bibinfo{author}{Meng, X.}, \bibinfo{author}{Li, Z.}, \bibinfo{author}{Zhang,
  D.}, \& \bibinfo{author}{Karniadakis, G.~E.} (\bibinfo{year}{2020}).
\newblock \bibinfo{title}{{PPINN}: Parareal physics-informed neural network for
  time-dependent {PDE}s}.
\newblock {\it \bibinfo{journal}{Computer Methods in Applied Mechanics and
  Engineering}\/},  {\it \bibinfo{volume}{370}\/}, \bibinfo{pages}{113250}.
  \DOIprefix\doi{10.1016/j.cma.2020.113250}.
\bibitem[{Nascimento et~al.(2018)Nascimento, D\'orea \&
  Gonçalves}]{mpc_robotics}
\bibinfo{author}{Nascimento, T.~P.}, \bibinfo{author}{D\'orea, C. E.~T.}, \&
  \bibinfo{author}{Gonçalves, L. M.~G.} (\bibinfo{year}{2018}).
\newblock \bibinfo{title}{Nonlinear model predictive control for trajectory
  tracking of nonholonomic mobile robots: A modified approach}.
\newblock {\it \bibinfo{journal}{International Journal of Advanced Robotic
  Systems}\/},  {\it \bibinfo{volume}{15}\/}.
  \DOIprefix\doi{10.1177/1729881418760461}.
\bibitem[{Nelles(2001)}]{nl_sys_ident}
\bibinfo{author}{Nelles, O.} (\bibinfo{year}{2001}).
\newblock {\it \bibinfo{title}{Nonlinear System Identification: From Classical
  Approaches to Neural Networks and Fuzzy Models}\/}.
\newblock (\bibinfo{edition}{1st} ed.).
\newblock \bibinfo{address}{Berlin}: \bibinfo{publisher}{Springer}.
\bibitem[{Nocedal \& Wright(2006)}]{Nocedal2006}
\bibinfo{author}{Nocedal, J.}, \& \bibinfo{author}{Wright, S.~J.}
  (\bibinfo{year}{2006}).
\newblock {\it \bibinfo{title}{Numerical Optimization}\/}.
\newblock (\bibinfo{edition}{2nd} ed.).
\newblock \bibinfo{address}{New York, NY, USA}: \bibinfo{publisher}{Springer}.
\bibitem[{Normey-Rico \& Camacho(2007)}]{Normey-Rico2007}
\bibinfo{author}{Normey-Rico, J.~E.}, \& \bibinfo{author}{Camacho, E.~F.}
  (\bibinfo{year}{2007}).
\newblock {\it \bibinfo{title}{Control of Dead-time Processes}\/}.
\newblock \bibinfo{publisher}{Springer London}.
\newblock \DOIprefix\doi{10.1007/978-1-84628-829-6}.
\bibitem[{Ortega \& Camacho(1996)}]{Ortega1996}
\bibinfo{author}{Ortega, J.~G.}, \& \bibinfo{author}{Camacho, E.}
  (\bibinfo{year}{1996}).
\newblock \bibinfo{title}{Mobile robot navigation in a partially structured
  static environment, using neural predictive control}.
\newblock {\it \bibinfo{journal}{Control Engineering Practice}\/},  {\it
  \bibinfo{volume}{4}\/}, \bibinfo{pages}{1669--1679}.
\bibitem[{Pan \& Wang(2012)}]{echo_pred}
\bibinfo{author}{Pan, Y.}, \& \bibinfo{author}{Wang, J.}
  (\bibinfo{year}{2012}).
\newblock \bibinfo{title}{Model predictive control of unknown nonlinear
  dynamical systems based on recurrent neural networks}.
\newblock {\it \bibinfo{journal}{IEEE Transactions on Industrial
  Electronics}\/},  {\it \bibinfo{volume}{59}\/}, \bibinfo{pages}{3089--3101}.
\bibitem[{Pang \& Karniadakis(2020)}]{Pang2020}
\bibinfo{author}{Pang, G.}, \& \bibinfo{author}{Karniadakis, G.~E.}
  (\bibinfo{year}{2020}).
\newblock \bibinfo{title}{Physics-informed learning machines for partial
  differential equations: Gaussian processes versus neural networks}.
\newblock {\it \bibinfo{journal}{Kevrekidis P., Cuevas-Maraver J., Saxena A.
  (eds) Emerging Frontiers in Nonlinear Science. Nonlinear Systems and
  Complexity}\/},  {\it \bibinfo{volume}{32}\/}, \bibinfo{pages}{323--343}.
\bibitem[{Raissi et~al.(2017)Raissi, Perdikaris \& Karniadakis}]{Raissi2017}
\bibinfo{author}{Raissi, M.}, \bibinfo{author}{Perdikaris, P.}, \&
  \bibinfo{author}{Karniadakis, G.~E.} (\bibinfo{year}{2017}).
\newblock \bibinfo{title}{Physics informed deep learning ({P}art {I}):
  Data-driven solutions of nonlinear partial differential equations}.
\newblock {\it \bibinfo{journal}{arXiv preprint arXiv:1711.10561}\/}, .
\bibitem[{Raissi et~al.(2019)Raissi, Perdikaris \& Karniadakis}]{Raissi2019}
\bibinfo{author}{Raissi, M.}, \bibinfo{author}{Perdikaris, P.}, \&
  \bibinfo{author}{Karniadakis, G.~E.} (\bibinfo{year}{2019}).
\newblock \bibinfo{title}{Physics-informed neural networks: A deep learning
  framework for solving forward and inverse problems involving nonlinear
  partial differential equations}.
\newblock {\it \bibinfo{journal}{Journal of Computational Physics}\/},  {\it
  \bibinfo{volume}{378}\/}, \bibinfo{pages}{686--707}.
  \DOIprefix\doi{0.1016/j.jcp.2018.10.045}.
\bibitem[{Schultz \& Rideout(1961)}]{iae}
\bibinfo{author}{Schultz, W.~C.}, \& \bibinfo{author}{Rideout, V.~C.}
  (\bibinfo{year}{1961}).
\newblock \bibinfo{title}{Control system performance measures: Past, present,
  and future}.
\newblock {\it \bibinfo{journal}{IRE Transactions on Automatic Control}\/},
  {\it \bibinfo{volume}{AC-6}\/}, \bibinfo{pages}{22--35}.
  \DOIprefix\doi{10.1109/TAC.1961.6429306}.
\bibitem[{Sirignano \& Spiliopoulos(2018)}]{Sirignano2018}
\bibinfo{author}{Sirignano, J.}, \& \bibinfo{author}{Spiliopoulos, K.}
  (\bibinfo{year}{2018}).
\newblock \bibinfo{title}{{DGM}: A deep learning algorithm for solving partial
  differential equations}.
\newblock {\it \bibinfo{journal}{Journal of Computational Physics}\/},  {\it
  \bibinfo{volume}{375}\/}, \bibinfo{pages}{1339--1364}.
  \DOIprefix\doi{10.1016/j.jcp.2018.08.029}.
\bibitem[{Stinis(2020)}]{Stinis2020}
\bibinfo{author}{Stinis, P.} (\bibinfo{year}{2020}).
\newblock \bibinfo{title}{Enforcing constraints for time series prediction in
  supervised, unsupervised and reinforcement learning}.
\newblock In {\it \bibinfo{booktitle}{Proceedings of the {AAAI} 2020 Spring
  Symposium on Combining Artificial Intelligence and Machine Learning with
  Physical Sciences}\/}.
\newblock volume \bibinfo{volume}{2587}.
\newblock \URLprefix \url{http://ceur-ws.org/Vol-2587/article\_5.pdf}.
\bibitem[{Terzi et~al.(2020)Terzi, Bonetti, Saccani, Farina, Fagiano \&
  Scattolini}]{Terzi2020}
\bibinfo{author}{Terzi, E.}, \bibinfo{author}{Bonetti, T.},
  \bibinfo{author}{Saccani, D.}, \bibinfo{author}{Farina, M.},
  \bibinfo{author}{Fagiano, L.}, \& \bibinfo{author}{Scattolini, R.}
  (\bibinfo{year}{2020}).
\newblock \bibinfo{title}{Learning-based predictive control of the cooling
  system of a large business centre}.
\newblock {\it \bibinfo{journal}{Control Engineering Practice}\/},  {\it
  \bibinfo{volume}{97}\/}, \bibinfo{pages}{104348}.
  \DOIprefix\doi{https://doi.org/10.1016/j.conengprac.2020.104348}.
\bibitem[{W{\"{a}}chter \& Biegler(2006)}]{Wachter2006}
\bibinfo{author}{W{\"{a}}chter, A.}, \& \bibinfo{author}{Biegler, L.~T.}
  (\bibinfo{year}{2006}).
\newblock \bibinfo{title}{{On the implementation of an interior-point filter
  line-search algorithm for large-scale nonlinear programming}}.
\newblock {\it \bibinfo{journal}{Mathematical Programming}\/},  {\it
  \bibinfo{volume}{106}\/}, \bibinfo{pages}{25--57}.
  \DOIprefix\doi{10.1007/s10107-004-0559-y}.
\bibitem[{Wei et~al.(2021)Wei, Li \& Stojanovic}]{Wei2021}
\bibinfo{author}{Wei, T.}, \bibinfo{author}{Li, X.}, \&
  \bibinfo{author}{Stojanovic, V.} (\bibinfo{year}{2021}).
\newblock \bibinfo{title}{Input-to-state stability of impulsive
  reaction{\textendash}diffusion neural networks with infinite distributed
  delays}.
\newblock {\it \bibinfo{journal}{Nonlinear Dynamics}\/},  {\it
  \bibinfo{volume}{103}\/}, \bibinfo{pages}{1733--1755}. \URLprefix
  \url{https://doi.org/10.1007/s11071-021-06208-6}.
  \DOIprefix\doi{10.1007/s11071-021-06208-6}.
\bibitem[{Witt \& Werner(2010)}]{APC}
\bibinfo{author}{Witt, J.}, \& \bibinfo{author}{Werner, H.}
  (\bibinfo{year}{2010}).
\newblock \bibinfo{title}{Approximate model predictive control for nonlinear
  multivariable systems}.
\newblock {\it \bibinfo{journal}{Model Predictive Control}\/},  (pp.
  \bibinfo{pages}{141--166}). \DOIprefix\doi{10.5772/46955}.
\bibitem[{Y.~Hafeez et~al.(2015)Y.~Hafeez, Ndikilar \& Isyaku}]{hafeez2015}
\bibinfo{author}{Y.~Hafeez, H.}, \bibinfo{author}{Ndikilar, C.~E.}, \&
  \bibinfo{author}{Isyaku, S.} (\bibinfo{year}{2015}).
\newblock \bibinfo{title}{Analytical study of the {Van} der {Pol} equation in
  the autonomous regime}.
\newblock {\it \bibinfo{journal}{Progress in Physics}\/},  {\it
  \bibinfo{volume}{11}\/}, \bibinfo{pages}{252--255}.
\bibitem[{Yang et~al.(2021)Yang, Meng \& Karniadakis}]{Yang2020}
\bibinfo{author}{Yang, L.}, \bibinfo{author}{Meng, X.}, \&
  \bibinfo{author}{Karniadakis, G.~E.} (\bibinfo{year}{2021}).
\newblock \bibinfo{title}{{B-PINNs}: Bayesian physics-informed neural networks
  for forward and inverse {PDE} problems with noisy data}.
\newblock {\it \bibinfo{journal}{Journal of Computational Physics}\/},  {\it
  \bibinfo{volume}{425}\/}, \bibinfo{pages}{109913}.
\bibitem[{Zhai \& Sands(2021)}]{Zhai2021}
\bibinfo{author}{Zhai, H.}, \& \bibinfo{author}{Sands, T.}
  (\bibinfo{year}{2021}).
\newblock \bibinfo{title}{Physics-informed deep operator control: Controlling
  chaos in van der pol oscillating circuits}.
\newblock {\it \bibinfo{journal}{arXiv preprint arXiv:2112.14707}\/}, .
\bibitem[{Zhu et~al.(2019)Zhu, Zabaras, Koutsourelakis \& Perdikaris}]{Zhu2019}
\bibinfo{author}{Zhu, Y.}, \bibinfo{author}{Zabaras, N.},
  \bibinfo{author}{Koutsourelakis, P.-S.}, \& \bibinfo{author}{Perdikaris, P.}
  (\bibinfo{year}{2019}).
\newblock \bibinfo{title}{Physics-constrained deep learning for
  high-dimensional surrogate modeling and uncertainty quantification without
  labeled data}.
\newblock {\it \bibinfo{journal}{Journal of Computational Physics}\/},  {\it
  \bibinfo{volume}{394}\/}, \bibinfo{pages}{56--81}.
  \DOIprefix\doi{10.1016/j.jcp.2019.05.024}.

\end{thebibliography}

\end{document}